# Bridging Brain with Foundation Models through Self-Supervised Learning

Hamdi Altaheri[1], Fakhri Karray[1,2], Md. Milon Islam[1], S M Taslim Uddin Raju[1], and Amir-Hossein Karimi[1]
[1]Department of Electrical and Computer Engineering, University of Waterloo, N2L 3G1, Ontario, Canada.
[2]Mohamed Bin Zayed University of Artificial Intelligence, Abu Dhabi, United Arab Emirates

***Abstract*—** Foundation models (FMs), powered by self-supervised learning (SSL), have redefined the capabilities of artificial intelligence, demonstrating exceptional performance in domains like natural language processing and computer vision. These advances present a transformative opportunity for brain signal analysis. Unlike traditional supervised learning, which is limited by the scarcity of labeled neural data, SSL offers a promising solution by enabling models to learn meaningful representations from unlabeled data. This is particularly valuable in addressing the unique challenges of brain signals, including high noise levels, inter-subject variability, and low signal-to-noise ratios. This survey systematically reviews the emerging field of bridging brain signals with foundation models through the innovative application of SSL. It explores key SSL techniques, the development of brain-specific foundation models, their adaptation to downstream tasks, and the integration of brain signals with other modalities in multimodal SSL frameworks. The review also covers commonly used evaluation metrics and benchmark datasets that support comparative analysis. Finally, it highlights key challenges and outlines future research directions. This work aims to provide researchers with a structured understanding of this rapidly evolving field and a roadmap for developing generalizable brain foundation models powered by self-supervision.

***Index Terms*—electroencephalogram (EEG), foundation models, Self-Supervised Learning, Generative AI, multimodal SSL, brain-computer interface (BCI), survey.**

## 1. Introduction

The past five years have witnessed a paradigm shift in machine learning, driven by self-supervised learning (SSL) and the emergence of foundation models–large models pre-trained on broad data. SSL has become a cornerstone of modern Artificial Intelligence (AI), drastically reducing the need for expensive labeled data and enabling more robust, scalable, and adaptable models [1]. Its evolution from early autoencoders [2], [3] and word2vec [4] to today's massive transformers [5] (and beyond) highlights a unifying principle: the structure in data itself is often enough to learn powerful and general representations. SSL can capture more knowledge about the underlying structure of the world compared to supervised or unsupervised learning, as it leverages virtually unlimited data and extracts extensive feedback from each example [1].

Foundation models, such as Large Language Models (LLMs), are a prime example of how SSL can lead to breakthroughs in AI models. By leveraging vast amounts of unlabeled text, LLMs learn rich semantic representations that generalize across diverse natural language processing (NLP) tasks, from machine translation to text generation. This success has inspired research beyond NLP, with SSL increasingly applied to domains such as computer vision, speech processing, and time-series data.

In the context of brain signal decoding, SSL offers the potential to extract meaningful representations from raw brain signals, mitigating the challenges of limited annotated data and inter-subject variability. As SSL continues to evolve, its impact on brain decoding could parallel the transformative role it has played in NLP and computer vision, paving the way for more data-efficient and generalizable brain-computer interface (BCI) systems.

Electroencephalography (EEG) is a widely employed neuroimaging technique renowned for its non-invasiveness and high temporal resolution in capturing the brain's electrical activity. Traditional machine learning (ML) and deep learning (DL) approaches have propelled EEG-based BCI toward improved performance in tasks such as motor imagery classification [6]–[8], cognitive workload monitoring, and emotion recognition. However, these methods often require extensive domain-specific feature engineering and are limited by data scarcity [9].

Bridging brain signals with foundation models through SSL is a promising strategy for addressing current limitations and offers unprecedented opportunities to advance our understanding of the brain. Although early results highlight the potential of this approach [10], integrating brain data with SSL into large-scale brain models remains an emerging field with numerous challenges. Unlike language models that operate in discrete text spaces with finite vocabularies, brain signals, such as EEG data, exist in a high-dimensional continuous space with complex temporal and spatial dynamics. Effectively modeling EEG requires capturing continuous fluctuations and low-amplitude signals, and addressing its inherently low signal-to-noise ratio (SNR), non-stationarity, and high inter-subject variability. Furthermore, building large-scale brain models involves aligning variability in channel configurations and acquisition protocols, especially when aggregating data from diverse sources.

In addition, training SSL models presents its own difficulties, including high computational costs associated with processing large-scale unlabeled data, a lack of detailed implementation guidance, the absence of a unified theoretical framework, and the complexity of selecting suitable pretext tasks and hyperparameters. Given these empirical challenges, entering the field of SSL for Brain signal decoding demands significant resources and experience.

In light of these considerations, this survey aims to lower the entry barrier for new researchers by providing practical guidelines, formalizations, taxonomies, and accessible insights into the application of SSL and foundation models in brain signal decoding, with a primary focus on EEG data.

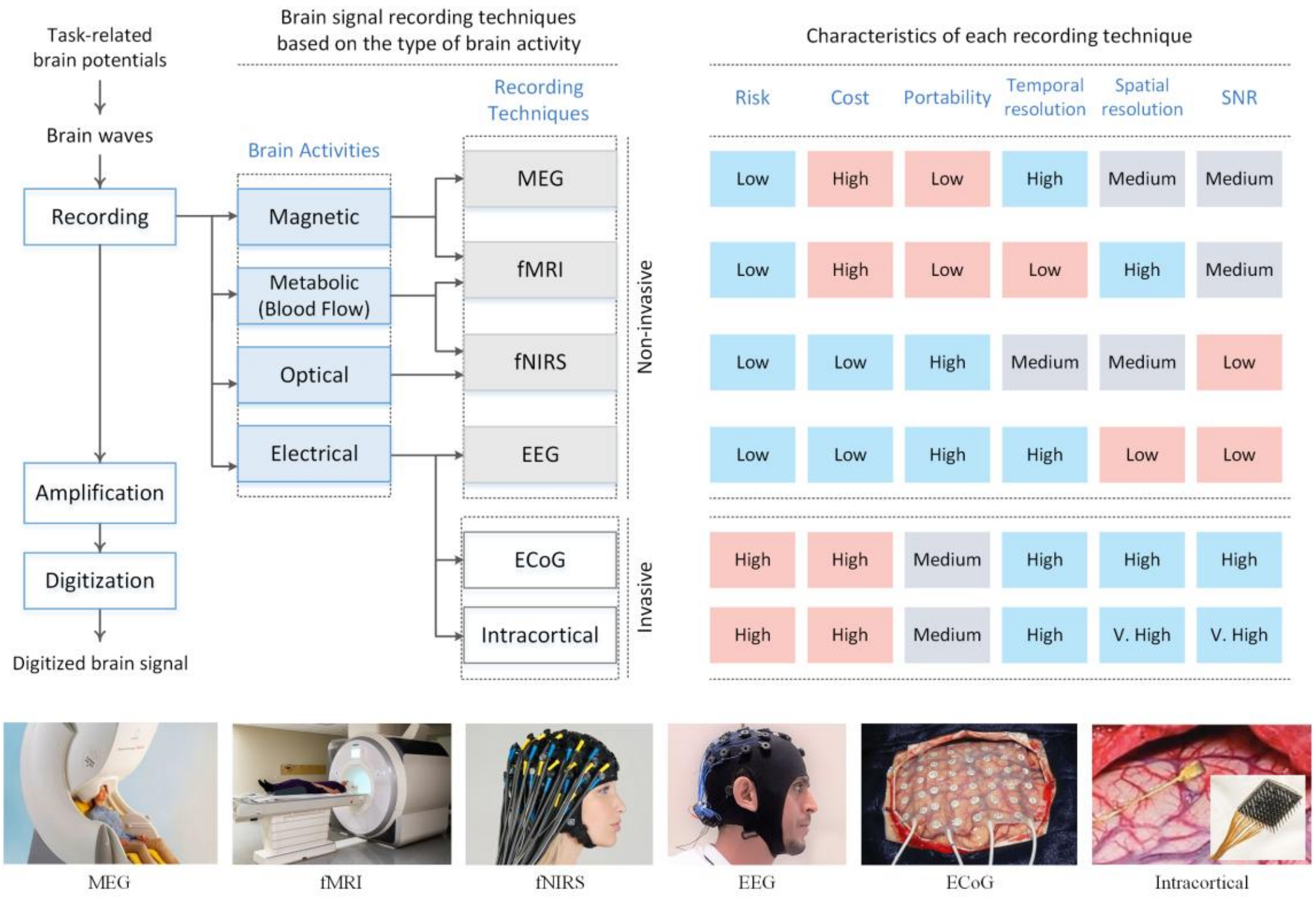

Figure 1. A taxonomy of brain signal acquisition showing different brain signal recording techniques and their characteristics.

This literature review systematically synthesizes the current body of knowledge on the application of self-supervised learning to bridge the gap between brain signals and foundation models. It explores the various SSL techniques employed, the adaptation of pre-trained models for downstream tasks, the current landscape of foundation models specifically developed for brain data, and the role of multimodal approaches. The review also outlines the evaluation strategies commonly used in the field, highlights relevant brain signal datasets, and finally, delves into the future directions and open challenges in this rapidly evolving area of research. While the primary focus is on EEG brain signals, the discussion section extends the scope to include brain decoding methods beyond EEG and their potential integration into future research.

Following this introduction, **Section 2** provides background on learning paradigms and introduces brain signal decoding with a focus on EEG. **Section 3** outlines the survey methodology, including databases, keywords, selection criteria, and timeline. **Section 4** reviews self-supervised learning (SSL) techniques in EEG, categorized into self-predictive and contrastive methods. **Section 5** surveys brain foundation models, including evaluation metrics and protocols. **Section 6** explores multimodal models integrating brain signals with other modalities. **Section 7** discusses adapting SSL models for classification and generative tasks and summarizes major EEG datasets for SSL. **Section 8** concludes with a discussion on integrating LLMs with brain signals, promising SSL strategies, practical guidelines, broader brain signal applications, limitations, and future research directions.

## 2. Background

This section provides the foundational context for the review by outlining key brain recording methods and core concepts related to self-supervised learning and foundation models. The aim is to clarify the conceptual and technical landscape in which this paper is situated and to clearly define the scope and focus of the review.

### 2.1 Brain signal recording techniques

Mental activity within the central nervous system (CNS) generates dynamic patterns known as brain waves or neural oscillations. These activities arise from neuronal communication, leading to changes in both ionic current flow and cerebral blood flow. Such physiological changes can be captured using various neuroimaging techniques. Brain electrical activity is measured using electrical signals (e.g., EEG) or magnetic fields (e.g., magnetoencephalography, MEG), while changes in cerebral blood flow are assessed using optical techniques (e.g., functional near-infrared spectroscopy, fNIRS) or magnetic imaging methods (e.g., functional magnetic resonance imaging, fMRI), as illustrated in Figure 1.

The choice of brain signal recording equipment depends on factors like application, cost, and user community. Based on the type of equipment used, BCI systems are generally classified into two categories: invasive and non-invasive. Invasive BCIs, such as ECoG and intracortical recordings, offer high spatial and temporal resolution but require surgery and costly equipment. Non-invasive methods, including EEG, MEG, fMRI, and fNIRS, measure brain activity externally. Among these, EEG is the most widely used due to its portability, affordability, ease of use, and high temporal resolution [11], as illustrated in Figure 1.

### EEG signals

EEG is a widely used non-invasive technique for recording electrical brain activity by detecting voltage fluctuations resulting from ionic currents within neurons [12]. However, since these signals must pass through the skull and other tissues before reaching the scalp electrodes, the recorded signal is significantly attenuated—capturing only about 5% of the original brain activity [13]. As a result, EEG signals typically require preprocessing to enhance signal quality before further analysis.

EEG data is generally represented as a two-dimensional matrix, where one axis corresponds to spatial information (electrode channels) and the other to temporal data (time points) [14]. Spatial resolution depends on the number and placement of electrodes, commonly ranging from 21 to 64 in research and clinical settings, although systems may support up to 256 electrodes. Temporal resolution is defined by the sampling rate, which typically ranges from 128 Hz to 1000 Hz. Figure 2 provides an example of a 23-channel EEG signal sampled at 256 Hz, and Figure 3 illustrates the standardized electrode placement on the scalp.

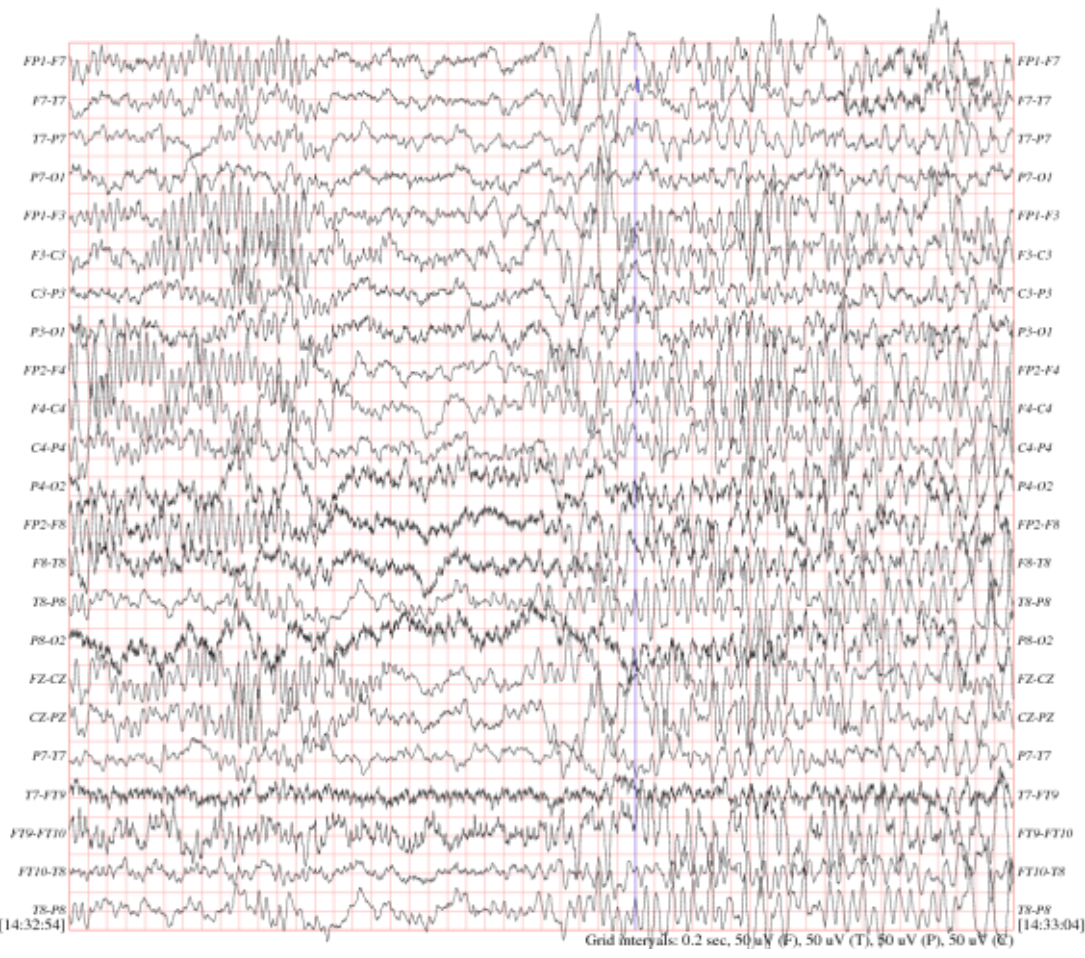

Figure 2. Example of a 23-channel EEG signal recorded over 10 milliseconds, sampled at 256 Hz with 16-bit resolution [15].

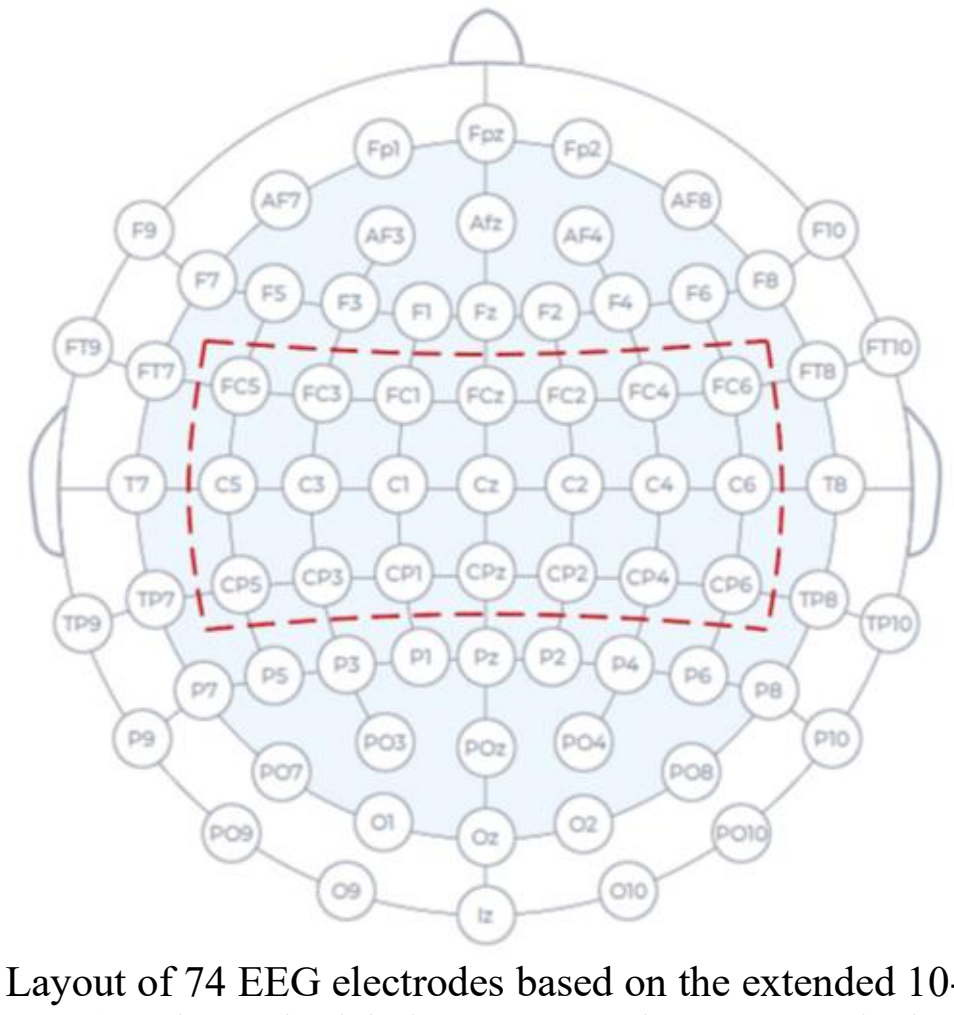

Figure 3. Layout of 74 EEG electrodes based on the extended 10-20 system (10-10 system). Electrode labels correspond to anatomical regions: C (central), T (temporal), F (frontal), Fp (prefrontal), P (parietal), and O (occipital). Intermediate positions are indicated by AF, FC, FT, CP, TP, and PO.

## 2.2 Conceptual foundations

To establish a clear context for this review, it is essential to define the relationships between key components of modern AI—self-supervised learning, foundation models, multimodal learning, and generative models—as illustrated in Figure 4.

### Self-supervised learning

SSL is a machine learning paradigm where a model learns from unlabeled data by generating their own surrogate labels or prediction tasks, known as pretext tasks. In other words, the data itself provides the supervision signal. For example, a language model might learn by predicting the next word in a sentence (as GPT models do during pre-training) without any human-provided labels. This approach leverages inherent structures in data (like parts of text or images) to train rich representations [1].

SSL methods generally fall into two broad categories: self-predictive learning and contrastive learning, as shown in Figure 4. In self-predictive learning, such as masked autoencoders [16], the model learns to reconstruct missing or corrupted parts of the input. In contrastive learning, the model learns to distinguish between different augmented views of the same input or between paired inputs from different modalities (e.g., an image and its caption). SSL has become a core training strategy for modern large-scale models, enabling them to learn effectively from vast amounts of unlabeled data [17].

### Foundation models

SSL has led to the rise of foundation models—large pre-trained architectures trained on broad, often unlabeled data from diverse sources [17]. These models are defined by their scale and generality: instead of building a separate model for each task, a single large model serves as a reusable backbone that can be fine-tuned or prompted to perform a wide range of downstream tasks.

### Multimodal learning

Many foundation models are extended through multimodal learning, allowing them to integrate and reason across multiple data types (e.g., vision and language), thus enhancing their contextual understanding and alignment across modalities.

### Generative models

The foundation models, whether unimodal or multimodal, also power generative models, which can produce new content such as text, images, or audio. Generative applications like ChatGPT (text generation) and DALL·E (image generation) build on foundation models to achieve their remarkable results.

This review focuses on studies that leverage unlabeled brain data using SSL to build large-scale foundation models.

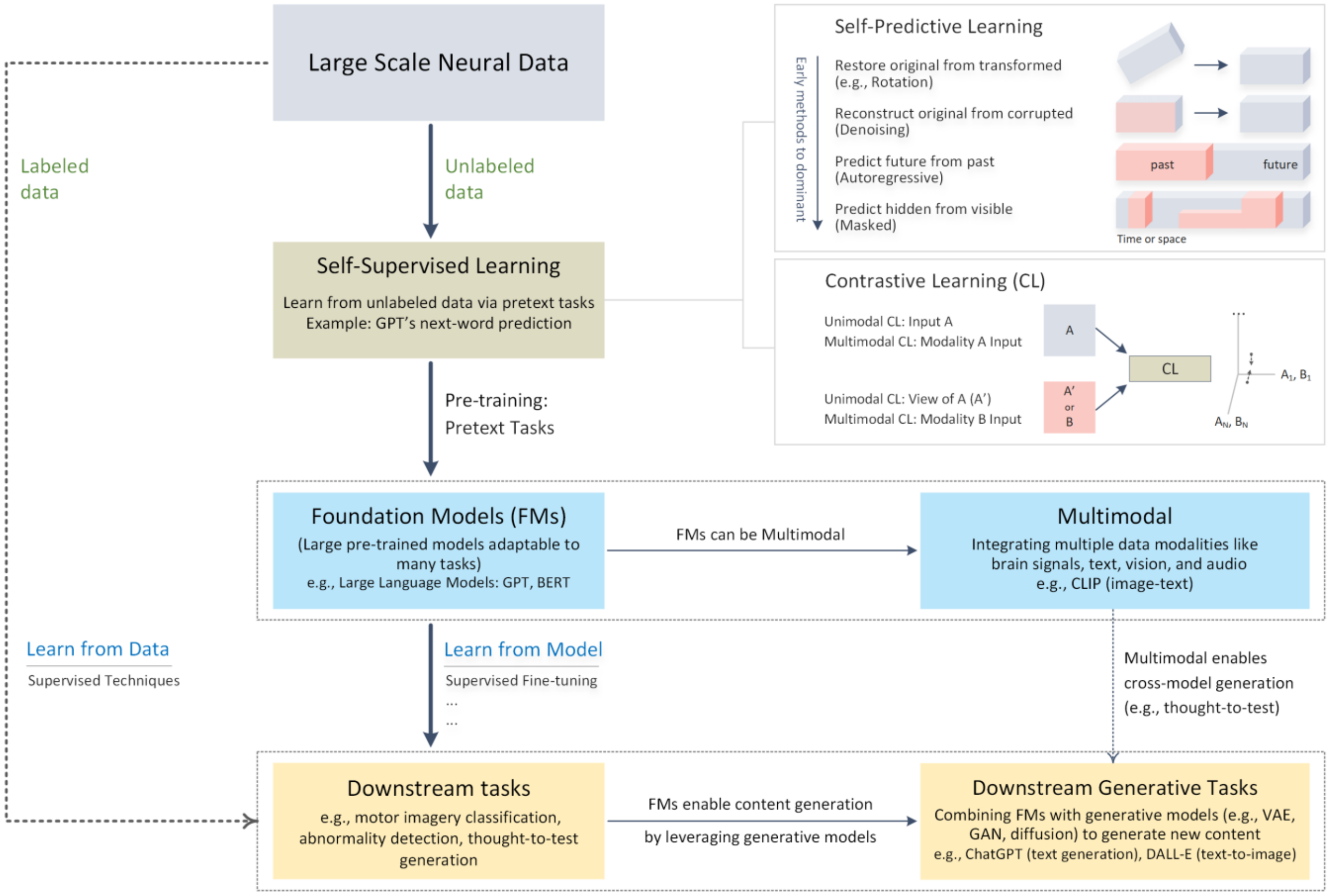

Figure 4. Conceptual diagram of core components in modern AI and their relationships. Self-supervised learning enables the training of foundation models from unlabelled data. Foundation models can be extended through multimodal learning to integrate multiple data types. Both unimodal and multimodal foundation models serve as the backbone for generative models that can produce text, images, or audio—including cross-modal outputs such as text-to-image generation.

It includes both unimodal (brain-only) and multimodal approaches that involve SSL applied to brain signals for either classification or generative tasks. Methods that do not utilize unlabeled brain data or omit SSL in multimodal setups are excluded from this review. We consider generative models only within the context of SSL, where generation tasks are guided by representations learned from unlabeled brain data. Generative approaches that aim to synthesize data from brain signals by aligning brain signals with pre-trained generative models—without applying SSL directly to brain data—fall outside the scope of this survey. Interested readers may refer to [18] for a detailed review of brain-conditional multimodal synthesis.

## 2.3 Notations and training paradigms

Given a brain signal $X \in \mathbb{R}^{C \times T}$, where $C$ is the number of channels and $T$ the total number of time points, the primary goal of self-supervised learning is to learn a function $F(x) \rightarrow \mathbb{R}^d$ that maps input data $X$ into a meaningful $d$-dimensional representation space, capturing essential features without relying on external labels. This is achieved through pretext tasks $T = \{t_1, t_2, \ldots, t_n\}$, which define learning objectives that exploit the intrinsic structure of unlabeled data. These tasks generate pseudo-labels $Y_p$, serving as internal supervisory signals that guide representation learning.

SSL architectures often follow an encoder-decoder structure:

- **Encoder** $f_\theta$: extracts general-purpose representations.
- **Pretext decoder** $g_\delta^{pt}$: works with the encoder to solve pretext tasks during pretraining.
- **Downstream decoder** $g_\xi^{dt}$: used in the fine-tuning phase to adapt the model for specific **downstream tasks** $dt$ (e.g., motor imagery classification, emotion recognition), which require labeled data.

In some cases, the representations learned during pretraining can be directly applied to downstream tasks without additional training—this is referred to as zero-shot SSL. When fine-tuning is used, it enhances task-specific performance by adapting the general representations to the target application.

SSL generally follows three training modes:

### 1. Pretraining–fine-tuning paradigm

The encoder is trained on pretext tasks and later fine-tuned on downstream tasks. This two-stage approach allows the effective transfer of learned features:

$$\theta, \delta = arg \min_{\theta,\delta} \mathcal{L}_{pt}(g_\delta^{pt}(f_\theta(X)), Y_p)$$

$$\theta, \xi = arg \min_{\theta,\xi} \mathcal{L}_{dt}(g_\xi^{dt}(f_\theta(X)), Y)$$

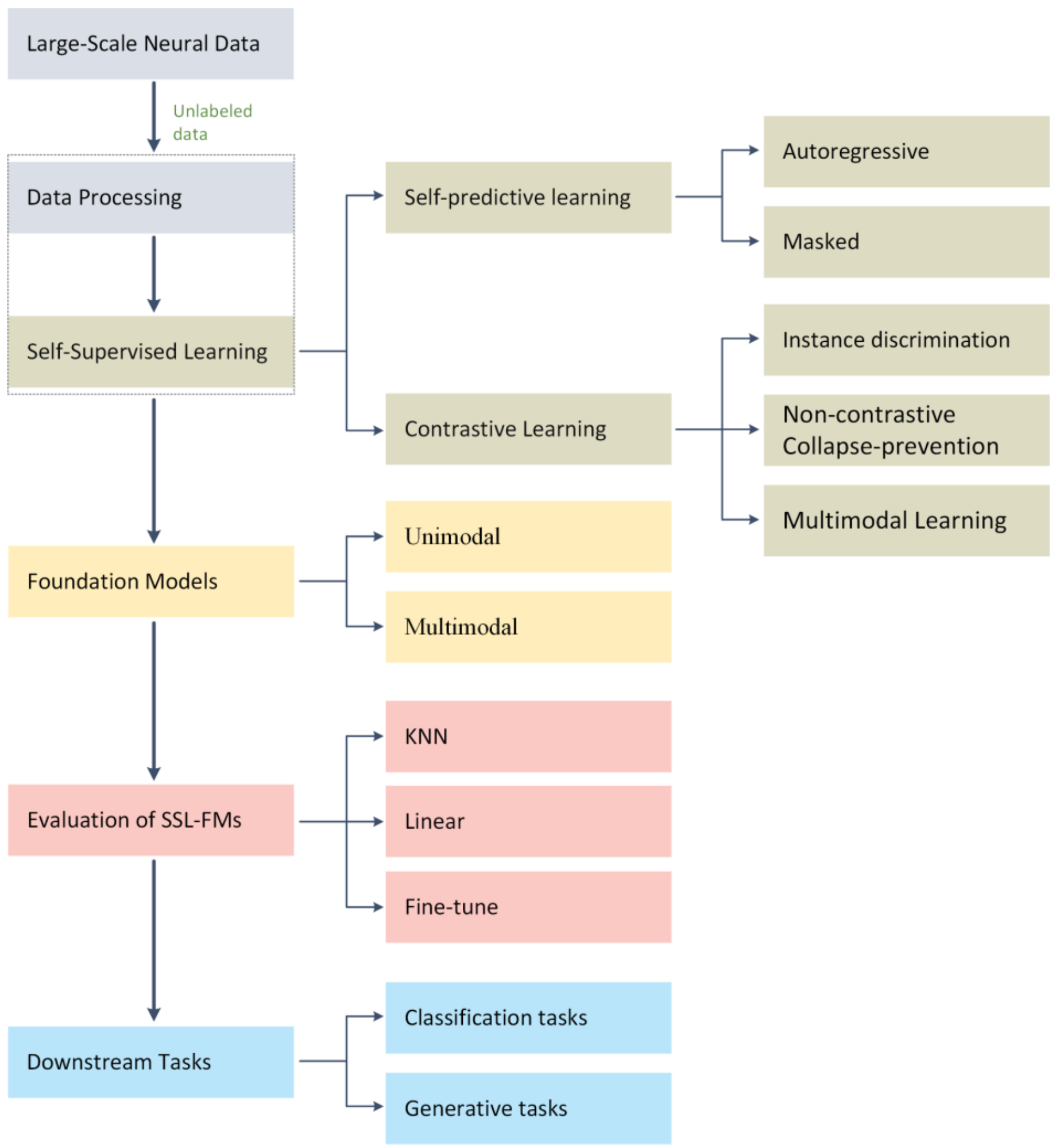

Figure 5. Overview of the core topics covered in this review. The figure illustrates the full pipeline for developing foundation models from large-scale neural data using self-supervised learning (SSL). It includes data processing, key SSL strategies (self-predictive and contrastive learning), and the construction of unimodal and multimodal models. These models are then evaluated using standard methods such as KNN, linear probing, and fine-tuning, before being applied to downstream tasks including classification and generative applications.

### 2. Joint-train mode

A combined loss function simultaneously optimizes both pretext and downstream tasks. This mode helps prevent overfitting and promotes more robust feature learning:

$$\theta, \delta, \xi = arg \min_{\theta,\delta,\xi} \alpha\mathcal{L}_{pt}( g_\delta^{pt}(f_\theta(X)), Y_p) + \beta\mathcal{L}_{dt}( g_\xi^{dt}(f_\theta(X)), Y)$$

Where α and β are loss-balancing hyperparameters.

### 3. Frozen encoder paradigm

Here, the encoder parameters are frozen after pretext training. Only the downstream decoder is fine-tuned to evaluate the quality of the learned representations:

$$\theta, \delta = arg \min_{\theta,\delta} \mathcal{L}_{pt}( g_\delta^{pt}(f_\theta(X)), Y_p)$$
$$\xi = arg \min_{\xi} \mathcal{L}_{dt}( g_\xi^{dt}(f_\theta(X)), Y)$$

Figure 5 provides an overview of the conceptual and methodological roadmap this survey follows in exploring the intersection of brain signal processing, self-supervised learning, and foundation models. Starting from unlabeled EEG datasets, we outline how SSL techniques—including self-predictive learning (e.g., autoregressive, masked modeling, and innate relationship learning) and contrastive learning (e.g., instance discrimination, non-contrastive methods, and multimodal learning)—are employed to build robust representations. These representations form the basis of unimodal or multimodal foundation models tailored for brain signal decoding. We then discuss common evaluation strategies (such as k-nearest neighbors, linear probing, and fine-tuning) to assess the quality of learned representations. Finally, we examine how these foundation models are adapted to downstream tasks, covering both classification and generative paradigms. This roadmap frames the structure of the survey and guides the reader through the key components and developments in this emerging field.

## 3. Method

To ensure a comprehensive and up-to-date review, we adopted a structured survey methodology following the PRISMA (Preferred Reporting Items for Systematic Reviews and Meta-Analyses) procedure [19], as shown in Figure 6. Using this procedure, three steps were performed sequentially. First, we identified relevant literature (last 10 years) from 2020 through early 2025 by querying two databases (Web of Science and PubMed). We searched using the following keywords (("Foundation Model*" OR "Large Language Model*" OR "Large Vision Model*" OR "Self-Supervised Learning" OR "Contrastive Learning") AND ("Brain Signal*" OR "Brain-Computer Interface" OR "Electroencephalogra*" OR "EEG" OR "Magnetoencephalogra*" OR "MEG" OR "Functional Magnetic Resonance Imaging" OR "fMRI" OR "Functional Near-Infrared Spectroscopy" OR "fNIRS" OR "Intracranial Electroencephalogra*" OR "iEEG" OR "Electrocorticogra*" OR "ECoG" OR "Intracortical")). We also traced references of key papers and included preprints (e.g. arXiv submissions) to capture very recent developments.. Duplicates between databases and unrelated studies were then screened out. After the papers were screened for relevance, full-text papers were assessed for eligibility according to the following constraints, which define the scope of the survey:

1. **Neuroimaging** —Exclude studies focusing solely on other bioimaging modalities without a brain component (e.g., ECG, EMG).
2. **Large self-supervised Models**—only research that uses self-supervised approaches on a large scale.
3. **Time**—This survey focuses on studies published within the past 10 years, with particular attention to developments following 2017, when self-supervised learning and large transformer-based architectures began to gain significant momentum.

Following the PRISMA procedure, 64 studies were selected for inclusion in this survey. The temporal distribution of these studies is illustrated in Figure 7. Although this review spans publications from the past decade, the application of SSL and foundation models to EEG decoding has only emerged within the last five years. The increase in publications after 2022 aligns with broader advancements in foundation models and multimodal AI, emphasizing the field's momentum and future potential.

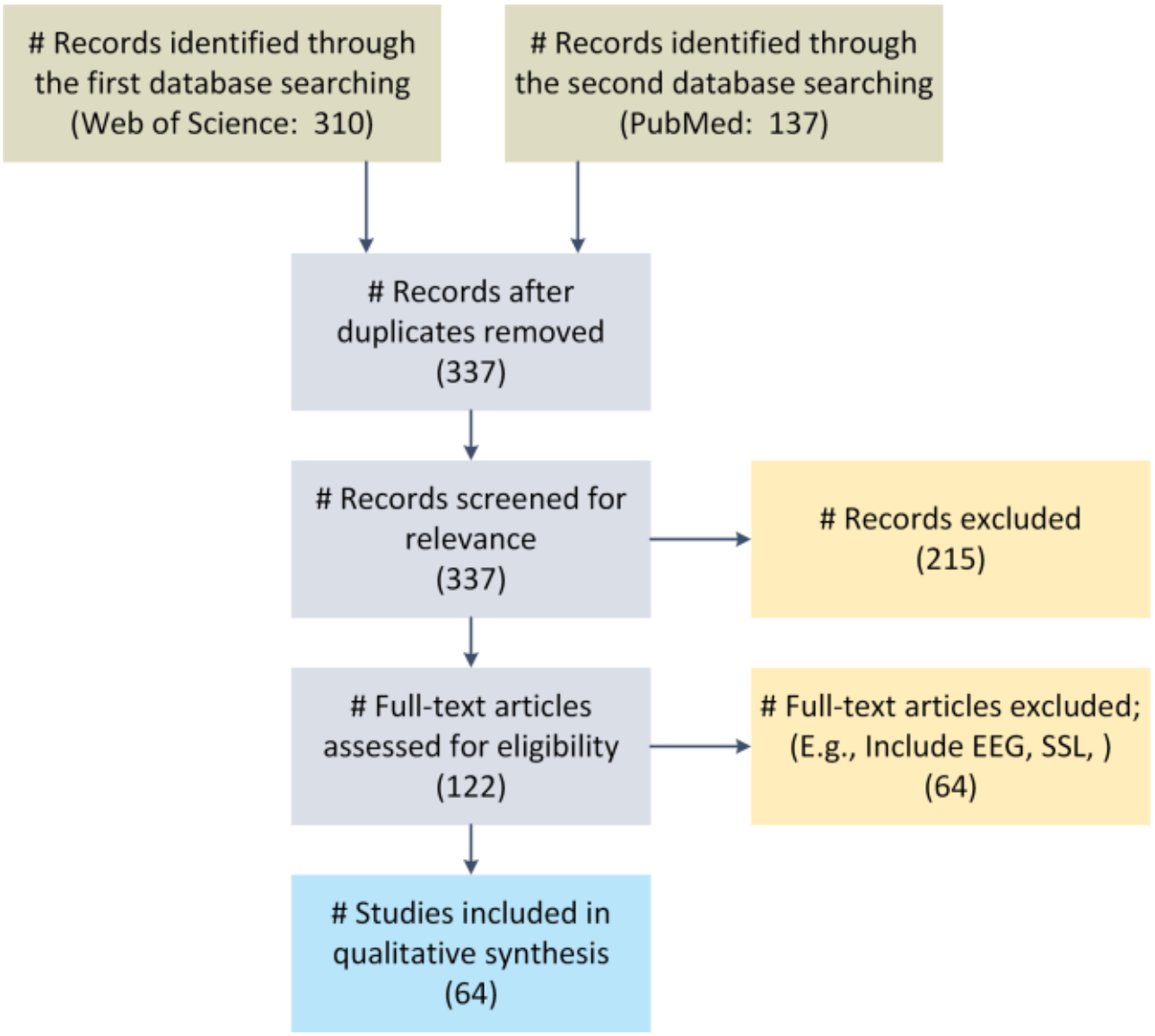

Figure 6. Diagram of article selection based on the PRISMA procedure.

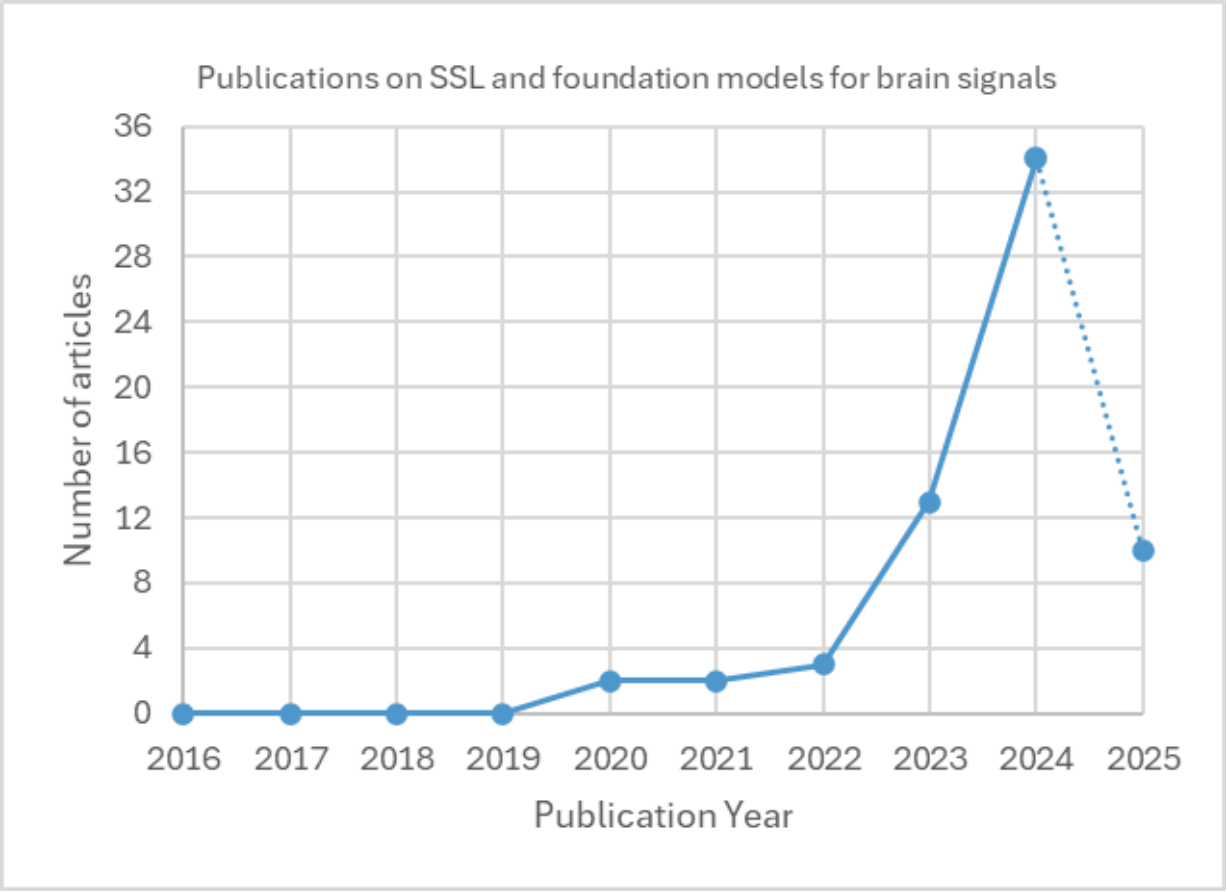

Figure 7. Number of articles applying self-supervised learning and foundation models and for EEG brain signal decoding in the past ten years.

## 4.1 Data extraction

From each included study, information on the following parameters was extracted:

- **Publication details**: Title, authors, venue, year.
- **The dataset used**: Type of EEG dataset (public vs. private), sample size, number of channels.
- **Architectural details**: Neural network architecture, inclusion of transformer-based or LLM-based components.
- **Evaluation metrics**: Accuracy, F1-score, mean squared error, etc.
- **Key findings**: Performance improvements, novel contributions, limitations.

The data was then synthesized to identify trends, gaps, and best practices for integrating foundation models with EEG signal analysis.

# 4. Self-supervised learning

Self-supervised learning is a machine learning technique that enables models to learn meaningful representations from unlabeled data by generating supervision signals from the data itself. The origins of SSL can be traced back to self-taught learning introduced by Raina et al. (2007) [20], which demonstrated how leveraging unlabeled data could enhance supervised learning performance. While earlier approaches focused on reconstruction-based unsupervised learning—such as denoising and variational autoencoders (2008-2013) [2], [3], [21]—SSL evolved into a distinct paradigm. [2], [3]

Initially, SSL emerged under the term unsupervised pre-training and gained significant attention with the introduction of word2vec [4] and GloVe [22] (2013-2014) in natural language processing. In computer vision, SSL expanded through pretext tasks such as relative patch prediction [23], image colorization [24], jigsaw puzzle solving [25], inpainting (filling the gaps) [26], and rotation prediction [27] between 2015 and 2018. The formalization of SSL as a key learning paradigm accelerated in the late 2010s with the rise of contrastive learning frameworks—such as MoCo [28], SimCLR [29], and BYOL [30] (2019–2020)—and masked modeling approaches like BERT [31] (2018) in NLP and masked autoencoders (MAE) [16] (2021) in computer vision. These advancements demonstrated SSL's effectiveness in learning rich representations from unlabeled data, making it a cornerstone of modern AI, particularly in NLP and computer vision applications.

Although SSL is sometimes used interchangeably with unsupervised learning, it is technically a subset of unsupervised learning, as it does not rely on human-provided labels. Unsupervised learning broadly encompasses techniques that operate without labeled data, such as traditional clustering and dimensionality reduction. SSL, however, occupies a unique position between supervised and unsupervised learning. It shares similarities with supervised learning because it optimizes a model against a predefined ground truth, which is derived automatically from the data rather than manually annotated. The supervision signal in SSL is not explicit labels, as in supervised learning, but pseudo-labels generated from the data by defining a learning objective based on pretext tasks, e.g., predicting a missing word in a sentence. This approach enables models to learn meaningful representations from unlabeled data while maintaining the structure and benefits of supervised training.

To effectively leverage large-scale unlabeled data, it is essential to define appropriate learning objectives that guide the model in learning meaningful representations. These objectives are typically implemented through carefully designed pretext tasks. For instance, in a pretext task where the model predicts missing parts of an image, the learning objective is to minimize the reconstruction error. Similarly, for a task that involves distinguishing between augmented views of the same image, the objective is to maximize the similarity between representations of those views. In this review, we use the terms SSL learning objectives and pretext tasks interchangeably to describe the pre-training phase of SSL models.

In NLP, SSL objectives often involve predicting missing or future words based on context. For example, BERT [31] is trained to predict masked words within a sentence, while GPT (GPT-1 [32] GPT-2 [33]) learns by predicting the next word. These tasks allow models to capture rich semantic and syntactic relationships from unlabeled text. The resulting representations are highly transferable and can be fine-tuned for various downstream applications such as machine translation, text summarization, and text generation.

In computer vision, similar SSL objectives have been developed. For example, Masked Autoencoders [16] train models to reconstruct missing patches of an image, encouraging a deep understanding of visual structures. Another common approach involves learning representations by maximizing the agreement between two augmented views of the same image, as implemented in methods like SimCLR [29], MoCo [28], and BYOL [30]. These approaches rely on contrastive or non-contrastive learning objectives that encourage the model to produce similar representations for different augmentations of the same input. Common augmentations include cropping, color jittering, flipping, and rotation—transformations that help the model learn representations that are invariant to such changes.

In the context of brain signals, defining effective pretext tasks and selecting appropriate learning objectives are critical to the success of self-supervised learning. Given the noisy and complex nature of brain signals, augmentations and task formulations must be tailored specifically to the signal's characteristics to ensure the model learns meaningful representations from large-scale unlabelled data.

The foundation of modern SSL is built on two key learning objectives: predicting missing or distorted parts of an input and contrasting different views of the same input. Based on these principles, SSL can be broadly categorized into two types: self-predictive learning and contrastive learning, as illustrated in Figure 8. Although various categorizations of SSL methods exist in the literature, often overlapping with one another, we find this division to be the most intuitive for helping readers grasp the fundamental principles of SSL. For a more in-depth discussion, including its theoretical foundations, readers may refer to [1].

Self-predictive learning methods involve restoring missing or altered parts of the input. This includes: (1) restoring the original signal from transformed versions (e.g., rotated or shuffled segments), (2) reconstructing corrupted input (denoising), (3) predicting future segments from past ones (autoregressive), and (4) predicting masked or hidden parts from visible ones (masked modeling). The vertical arrow indicates the historical evolution from early methods to more dominant strategies. Contrastive learning strategies rely on learning representations by comparing pairs of inputs. In unimodal contrastive learning, the model distinguishes between different augmented views of the same input (e.g., A and A′). In multimodal contrastive learning, inputs come from different modalities (e.g., A and B, such as EEG and text). Both approaches aim to align semantically similar representations while pushing apart dissimilar ones in the learned feature space.

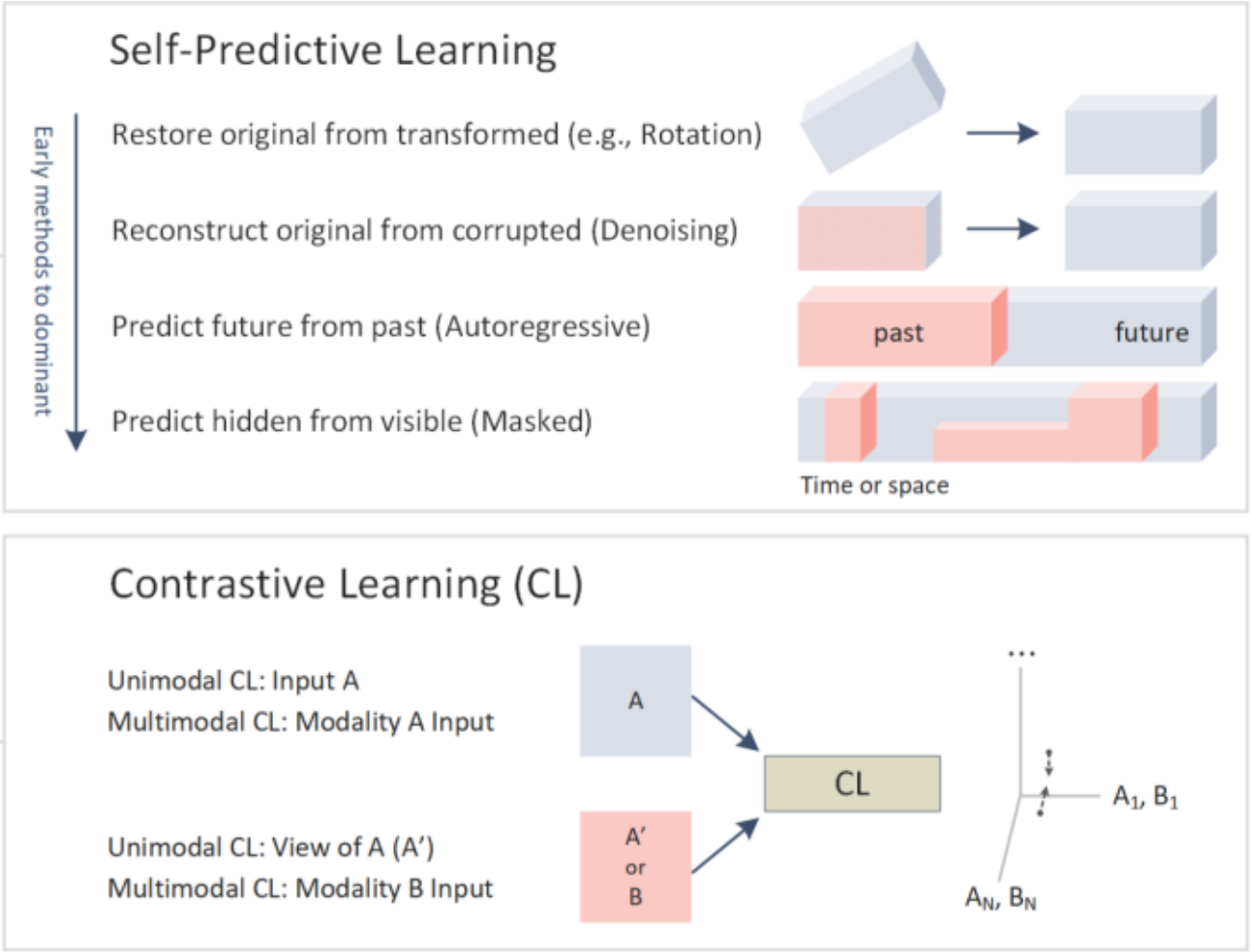

Figure 8. Overview of self-supervised learning strategies: (top) self-predictive learning methods, which reconstruct or predict parts of the input (e.g., masked, corrupted, or future data); (bottom) contrastive learning, which aligns representations of positive pairs (e.g., augmented views or cross-modal pairs) while separating negatives.

Before applying SSL strategies, it is essential to preprocess neural data to address the variability and inconsistencies commonly found in large-scale neural datasets. This includes differences in the number and placement of channels, as well as variations in signal length and sampling rates in different datasets. Data processing ensures that input signals are standardized, aligned, and formatted to be compatible with model training and evaluation. The following sections begin with an overview of data processing techniques, followed by a detailed exploration of SSL strategies.

## 4.1 Data processing: standardizing neural data for foundation models

Data processing is a critical component in building brain foundation models, as it ensures consistency and compatibility across large-scale, heterogeneous neural datasets. This step addresses two primary challenges: inconsistent spatial configurations (channel layouts) and variable temporal structures (signal lengths). Effectively addressing these is paramount for building robust and adaptable BFMs.

**Spatial heterogeneity: variable channel configurations**

One of the key issues in Brain signal data arises from the lack of standardization in electrode or sensor placements across different datasets. Brain signal data, whether from EEG, iEEG, fMRI, or other modalities, exhibits substantial spatial inconsistencies. Each "channel" (representing an electrode, sensor, or brain region voxel) captures neural activity from a specific location, but the number, spatial arrangement, and recording characteristics of these channels often vary significantly among datasets. For instance, the SEED [34] dataset employs a Neuroscan 62-channel configuration, while the DEAP dataset [35] utilizes a Biosemi 32-channel EEG system, leading to mismatches in both spatial coverage and resolution. Since each channel captures potentially unique neurophysiological information tied to its location, simply interpolating missing channels or resampling to a common configuration risk distorting critical spatial patterns. This poses significant challenges for integrating multiple datasets into a unified training pipeline for foundation models.

**Temporal heterogeneity: variable signal lengths**

Another major challenge involves temporal inconsistencies due to differing experimental paradigms. Brain signal recording durations vary based on the task: motor imagery may require only a few seconds, steady-state visual evoked potentials (SSVEP) only milliseconds, while emotion recognition or sleep studies may involve minutes to hours of recording. Such variability in temporal resolution introduces difficulties in standardizing data input for model training and evaluation.

A one-size-fits-all approach to signal duration is insufficient. Using fixed, arbitrary time windows can be detrimental: overly long windows may introduce irrelevant noise and obscure short-term events, while excessively short windows may miss slower, sustained neural dynamics, leading to a loss of important contextual information. This variability underscores the need for preprocessing methods that can unify or adapt to diverse temporal structures without losing essential task-related signals.

**Discretized signal segmentation**

To address structural and temporal inconsistencies in brain signals, a widely adopted strategy is discretized segmentation. This method partitions continuous raw brain signals—represented as $X \in \mathbb{R}^{C \times T}$, where $C$ is channels and $T$ is time steps—into standardized, fixed-length segments or patches. A sliding window approach is commonly employed, defined by a window length $w$ and a stride $s$. This generates temporal segments $x_{c,k}$ for each channel c at time step $k$, as formalized by:

$$x = \left\{ x_{c,k} \in \mathbb{R}^{w} \mid c = 1,2,\ldots C;\ k = 1,2,\ldots, \left\lfloor \frac{T-w}{s} \right\rfloor \right\}$$

The total number of generated patches is given by:

$$|x| = C \times \left\lfloor \frac{T-w}{s} \right\rfloor + 1$$

This segmentation approach has two main benefits: channel-wise consistency, where each channel is partitioned independently, supporting flexible processing of diverse channel counts. Temporal unification, where fixed-length segments standardize signal durations for subsequent modeling. This strategy imposes a uniform input structure, allowing models like Transformers to process data from varied sources by treating each segment as a token. It handles channel variability by processing each channel's stream independently and unifies signal durations by breaking them into standard chunks.

While discrete segmentation standardizing format, segmentation inherently strips away the original spatiotemporal context of each signal fragment. The model initially doesn't know where in the brain (channel/region) or when in the continuous recording a segment originated. This loss makes it difficult to model long-range temporal dependencies or complex spatial interactions, such as the synchronized activity between distinct brain areas (e.g., temporal and parietal lobes during audiovisual integration [36]).

**Positional encoding**

To address this limitation, positional encoding (PE) is introduced to convey the relative positions of segments in both the channel (spatial) and time (temporal) dimensions. Current foundation models utilize two primary PE strategies: Fixed Positional Encoding and Learnable Positional Encoding. Fixed positional encoding employs pre-defined mathematical functions (e.g., sinusoidal functions) to encode positions. These encodings are not learned during training. For instance, Brain-JEPA [37] uses a PE matrix based on brain gradient positioning for spatial information and sinusoidal functions for temporal information. This approach is computationally simple but less flexible. The learnable positional encoding uses trainable parameters that are optimized during model training to adaptively capture the dataset-specific spatiotemporal properties. For example, Brant [38] adds a learnable matrix $W_{pos}$ to projected input segments, while LaBraM [10] further divides positional encoding into temporal and spatial matrices $W_{\text{temp}}$ and $W_{\text{spat}}$. This offers greater flexibility but requires sufficient data to learn meaningful encodings.

**Limitations of current strategies**

Despite early successes, current data processing methods in brain foundation models often lack the flexibility and adaptability required for diverse neural tasks.

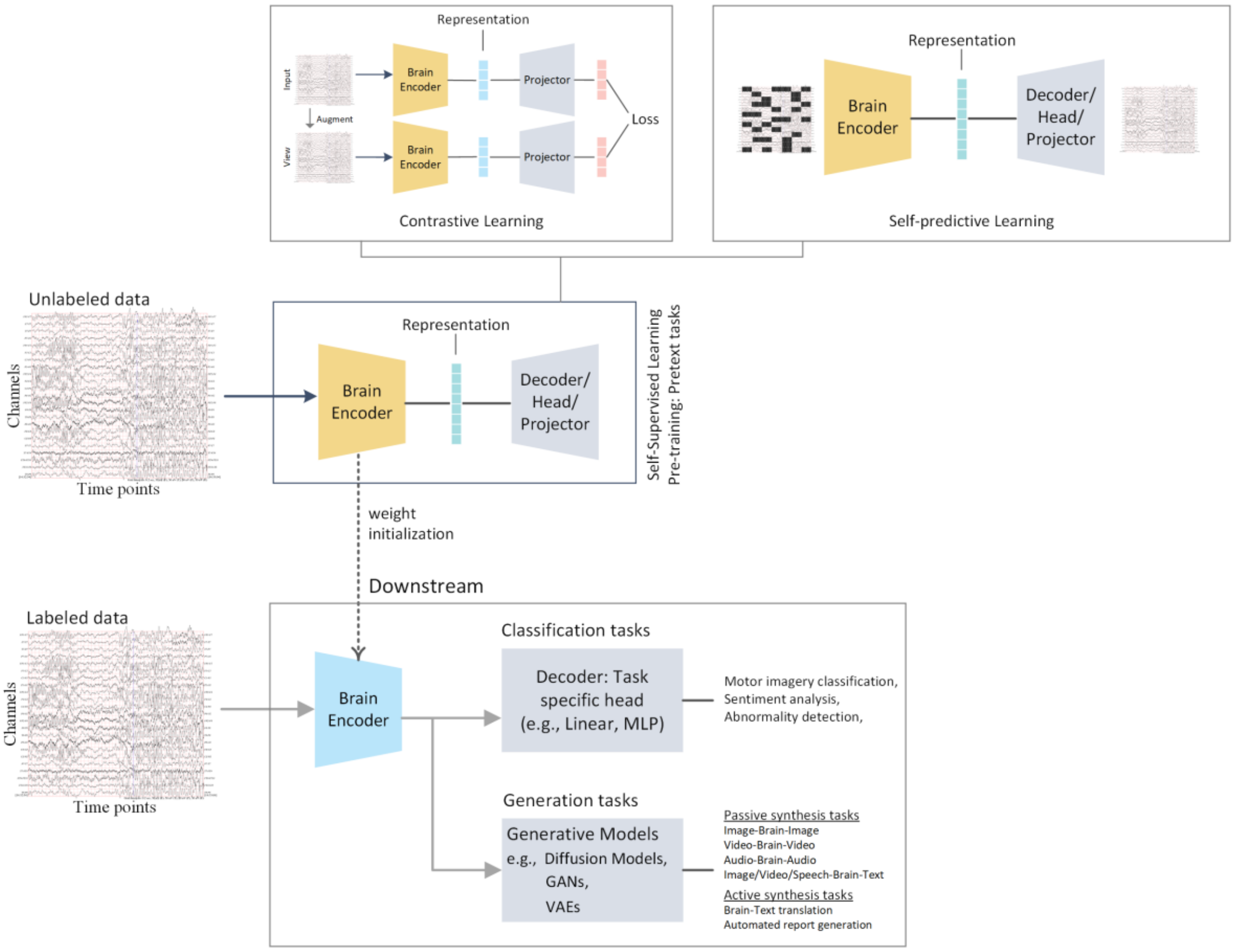

Figure 9. Overview of self-supervised learning (SSL) strategies for brain signal decoding. The top panel illustrates two major SSL paradigms: contrastive learning, which aligns representations of augmented views using a contrastive loss, and self-predictive learning, which reconstructs masked or transformed portions of the input. The middle panel shows the general SSL pretraining framework using unlabeled EEG/MEG data to learn representations via a brain encoder. These representations are then transferred to downstream tasks through weight initialization. The bottom panel outlines downstream applications, including classification tasks (e.g., motor imagery, sentiment analysis) and generation tasks (e.g., brain-to-text or brain-to-image synthesis), utilizing either task-specific heads or generative models.

**Fixed window constraints**—where segment lengths are predetermined—can fail to capture task-specific dynamics (e.g., SSVEP versus sleep studies) and may overlook inter-individual variability in neural response latencies. Short windows risk missing sustained cognitive states, while long windows can obscure transient patterns and introduce noise. This fundamental mismatch between static segmentation and inherently dynamic brain activity introduces bias and limits the generalizability of brain foundation models.

**Inconsistent coverage of brain channels/regions**: current methods for addressing variability in channel/region coverage across datasets often rely on simplistic compromises. A common strategy is to take either the intersection of shared channels—as in CBraMod [39], which ensures compatibility but discards potentially valuable data from non-overlapping regions—or the union of all recorded channels—as in EEGPT [40], which preserves more data but increases computational cost and requires additional handling of missing inputs. This includes techniques like imputation (estimating missing values from neighboring channels) or masking (flagging missing channels for the model). Neither approach fully captures the biological complexity or structural variability of brain signals, limiting the model's capacity to represent functional connectivity or adapt to diverse datasets.

**Loss of context**: Spatial and temporal segmentation can disrupt critical dependencies across channels and time steps by treating brain regions as isolated, independent units during standardization. This overlooks the structural and functional relationships between brain areas, leading to the underrepresentation of inter-regional connectivity and neural topology. As a result, the learned features may lack biological relevance. More flexible methods—such as graph-based models—are needed to better capture the underlying neurophysiological structure.

## 4.2 Self-predictive learning

Self-predictive learning, also known as auto-associative learning, involves models that predict missing or distorted segments of input data by leveraging the available information within the same sample. In this approach, the model artificially generates unknown portions or structures and subsequently attempts to reconstruct or predict them. Models trained using these methods are generally generative rather than discriminative, focusing on capturing inter-feature dependencies within a sample (intra-sample relationships).

Self-predictive techniques in SSL encompass a range of pretext tasks designed to learn meaningful representations from unlabeled data, as illustrated in Figure 8. Early methods focused on innate relationship prediction tasks, such as rotation prediction [27], where the model learns to classify the degree of rotation applied to an input. Self-predictive techniques also include traditional autoencoders, such as denoising autoencoders [2] and variational autoencoders [21], which reconstruct inputs from compressed latent representations. More recently, masked modeling and autoregressive strategies have emerged as dominant paradigms. Masked modeling—exemplified by masked autoencoders [14] in computer vision and masked language modeling in NLP (e.g., BERT [31])—trains models to predict missing parts of the input based on the surrounding context. In contrast, autoregressive approaches—such as causal language modeling (e.g., GPT [32])—learn to predict future elements in a sequence from past observations.

In the domain of SSL for neural data, the two most widely adopted self-predictive strategies are masked modeling and autoregressive learning. These approaches closely parallel the concepts of masked token prediction and next-token prediction used in LLMs. By treating segments of neural data—such as EEG chunks—as tokens, these methods can enable effective pretraining and contribute to the development of generalizable and robust brain foundation models. The following sections provide a detailed discussion of these techniques within the context of neural data.

## Masked modeling techniques

Masked modeling techniques are widely used in natural language processing [31] and computer vision [16]. In this technique, portions of an unlabeled data sample are masked, and the model is trained to predict or reconstruct the missing information. The original unmasked input serves as the ground truth, guiding the learning process through a reconstruction loss function.

Masked modeling—such as masked autoencoders—is sometimes considered a subset of denoising autoencoders, where the "noise" is interpreted as masked portions of the input. However, unlike denoising autoencoders, which aim to remove unwanted noise, masked autoencoders are designed to restore removed information.

This approach has proven effective in learning meaningful contextual representations for downstream tasks, typically using a high masking ratio for visual signals [16], [41] and a low masking ratio for natural language [31], [33]. In neural signals, masking can be applied to time segments or frequency bands, encouraging the model to learn more generalized and robust features from the data.

Masked prediction enables models to learn robust global representations from incomplete input data. From a neuroscience perspective, the effectiveness of masked prediction resonates strongly with the principle of redundant coding observed within the brain itself. Neural information is often distributed and overlapping, meaning activity in certain areas or at specific times can frequently be inferred or predicted from related activity elsewhere. For example, EEG, like other time-series signals, inherently exhibits strong temporal correlations, where activity at one moment is often predictive of the activity immediately following it. Masked prediction leverages this by forcing the model to learn these temporal dynamics to fill in missing time segments. The brain also employs distributed representation mechanisms where multiple regions often work in concert, collaboratively encoding the same cognitive states, sensory information, or motor intentions. This functional connectivity implies that activity in one region may be partially predictable from activity in functionally related regions. Masking channels or spatial regions encourages the model to learn these inter-regional dependencies.

The masked modeling approaches in brain signal processing can be categorized and compared as follows:

**Vector quantization-based approaches**: Chen et al., [42] and Jiang et al., [10] both utilize vector quantization with masked prediction, but differ in their implementation. Chen's EEGFormer focuses on creating interpretable discrete tokens through vector quantization, making the model more transparent but potentially losing fine-grained information. In contrast, Jiang's LaBraM combines vector quantization with neural spectrum prediction, offering a more comprehensive approach to temporal-frequency relationships but requiring more computational resources.

**Graph-enhanced masked modeling:** Wang et al., [43] combined masked autoencoding with graph neural networks. This method explicitly models inter-channel relationships in EEG data, providing better spatial awareness compared to pure transformer approaches. However, this architectural complexity makes it more computationally intensive than simpler masked modeling approaches.

**Knowledge-guided masking:** Kommineni et al., [46] takes incorporated domain-specific knowledge into the masked prediction framework using state space-based architecture (S4). This method differs from other approaches by explicitly incorporating physiological constraints, potentially improving biological plausibility but limiting flexibility compared to more general approaches.

**Fourier-based masked modeling:** Wu et al., [50] implement masking in both spatiotemporal and Fourier domains, distinguishing themselves from approaches that work solely in the time domain. This dual-domain approach offers better frequency-aware representations but may be more sensitive to noise compared to time-domain-only approaches. BrainBERT [51] implements a time-frequency masking strategy on iEEG spectrograms derived via the short-time Fourier transform or Superlet transform. In this setup, randomly selected spectral segments are masked, and a Transformer-based model—leveraging self-attention mechanisms—is trained to reconstruct the missing components using contextual cues from the unmasked regions. Brain-JEPA [37] introduces another spatiotemporal joint masking strategy, particularly suited for fMRI data. Unlike BrainBERT, which focuses on masking discrete time-frequency elements, Brain-JEPA applies masking across both spatial (brain regions) and temporal dimensions simultaneously. It employs cross-region, cross-time, and dual masking techniques, making the learning process more robust to diverse occlusion patterns. Furthermore, instead of reconstructing raw signals, Brain-JEPA predicts latent representations within an abstract feature space. This architectural choice reduces the impact of noise inherent in raw neuroimaging data and enhances the stability and transferability of the learned features across downstream tasks.

Table 1. Summary of studies using SSL self-predictive methods (autoregressive and masked prediction) applied to EEG brain signals.

| Study | SSL Method | Key Features | Performance Metrics | Modality Support |
|---|---|---|---|---|
| Wang et al., 2025 [39] | Patch-based masked | Criss-cross transformer to separately process spatial and temporal EEG patterns | Accuracy, Cohen's Kappa, Weighted F1 | EEG |
| Jiang et al., 2025 [44] | Autoregressive, Masked | Vector-quantized temporal-frequency prediction, multi-channel autoregression | Balanced Accuracy, Cohen's Kappa, Weighted F1, AUC-PR, AUROC | EEG |
| Wang et al., 2024 [43] | Masked | Graph-Enhanced EEG Foundation Model | Accuracy, AUROC | EEG |
| Fang et al., 2024 [45] | Masked | Cross-modal reconstruction objectives | No mention found | Multiple (not specified) |
| Chen et al., 2024 [42] | Masked | Vector-quantized Transformer model | AUROC, AUPRC, M-AUROC, M-AUPRC | EEG |
| Jiang et al., 2024 [10] | Masked | Masked EEG modeling with vector-quantized neural spectrum prediction | Balanced Accuracy, AUC-PR, AUROC, Cohen's Kappa, Weighted F1 | EEG |
| Kommineni et al., 2024 [46] | Masked | The knowledge-guided objective with S4 architecture | Accuracy | EEG |
| Cui et al., 2024 [47] | Autoregressive | Masked reconstruction using the GPT model | Classification accuracy | EEG |
| Yue et al., 2024 [48] | Autoregressive | Autoregressive EEG pre-trained model | Accuracy improvements over specialist models | EEG |
| Wang et al., 2024 [49] | Masked Prediction and Contrastive learning (Multi-modal Learning) | Contrastive EEG-Text Masked Autoencoder | BLEU and ROUGE-1 scores | EEG, Text |
| Wu et al., 2022 [50] | Masked | Fourier-based modeling framework | No mention found | EEG, EMG |

**Cross-modal masked approaches:** Wang et al., [49] and Fang et al., [45] extend masked modeling to multi-modal scenarios. Wang's Contrastive EEG-Text Masked Autoencoder combines masked prediction with contrastive learning, while Fang focuses on cross-modal reconstruction objectives. These approaches offer richer representations through modal alignment but face additional challenges in balancing information from different modalities.

**Patch-based masking:** Wang et al., [39] "CBraMod" employs a distinctive patch-based masked EEG reconstruction using a criss-cross transformer, separating spatial and temporal dependency modeling. This approach offers more structured feature learning compared to random masking strategies but may miss certain global patterns.

The choice between these methods often depends on the specific application requirements, with trade-offs between computational efficiency, biological plausibility, and representation richness. Multi-modal approaches show promising results but require more resources, while simpler masked modeling approaches offer better efficiency but potentially less rich representations.

## Autoregressive models

Autoregression (AR) is a technique used for modeling time-series data, where predictions are made based on past values of the same variable. Unlike conventional regression, e.g., linear regression, which uses external independent variables to predict a dependent variable, autoregression treats the variable itself as both the independent and dependent variable. This is why it is called "auto" regression—the model learns patterns within the data's own historical values. In an autoregressive model, the value at a given time step ($X_t$) is predicted using a weighted combination of its previous values ($X_{t-1}$, $X_{t-2}$,…, $X_{t-p}$), where $p$ is the order of the model. A simple AR model of order $p$, $AE(p)$, is expressed as:

$$X_t = \sum_{i=1}^{p} \varphi_i X_{t-\mathrm{i}} + \varepsilon_t$$

Where: $X_t$ is the value at time t, $\varphi_1, \varphi_2, \dots, \varphi_p$ are the model's learned parameters, and $\varepsilon_t$ is the error term (white noise).

Autoregressive models form the backbone of causal language modeling, where the goal is to predict the next token in a sequence based on the previous context. This approach underpins prominent foundation models such as GPT [32], LLaMA [ref], and Claude [ref], which have demonstrated strong performance in tasks like text generation, question answering, and dialogue systems.

Autoregressive models learn temporal representations by predicting future segments of neural signals using only past observations. This approach aligns with the brain's temporal integration processes, where current neural states are influenced by preceding activity due to mechanisms such as synaptic transmission and inter-regional signal propagation delays. By processing information sequentially and unidirectionally, AR models are naturally suited to capture the dynamic evolution of brain activity and learn long-range temporal dependencies. Unlike masked prediction methods, which leverage bidirectional context (information both before and after a masked segment) to learn representations, AR models strictly adhere to temporal causality. This makes them particularly relevant for modeling the directional flow of information within brain networks and for generative tasks involving sequential prediction.

For instance, Neuro-GPT [47] exemplifies this approach for EEG data. It employs a Transformer architecture, similar to GPT-2 [33] in NLP, to autoregressively predict subsequent EEG signal segments based on the preceding sequence. This training forces the model to internalize the temporal patterns characteristic of EEG signals. Similarly, Yue et al. [48] present EEGPT, a large-scale model with up to 1.1 billion parameters, which employs electrode-wise modeling and autoregressive pretraining. The model features a task-shared electrode graph network designed to support multi-task learning, enhancing versatility across different EEG tasks.

NeuroLM [44] utilizes an autoregressive framework with multimodal architecture that combines multi-channel autoregression with vector-quantized temporal-frequency prediction. By processing multi-channel EEG input and employing a causal, decoder-style structure, it predicts future EEG representations or associated linguistic tokens autoregressively, thereby learning to capture the temporal structure of EEG with high fidelity while potentially linking it to cognitive states expressed through language. The inherent causality enforced by the AR process enables these models to learn sequential dependencies and model the progression of neural states over time.

These models demonstrate the effectiveness of autoregressive approaches in modeling the temporal dynamics of neural signals, providing insights into the brain's functional organization.

## 4.3 Contrastive learning

## A. Concepts of contrastive learning

Contrastive Learning (Learning by Comparison) represents a distinct paradigm within SSL, shifting the focus from predicting parts of a single input (as in masked or autoregressive methods) to learning representations by comparing different data samples. The central premise is that meaningful features can be learned by understanding the similarities and differences between instances. CL methods operate by projecting input data into a lower-dimensional embedding space where representations of "similar" samples are pulled closer together, while representations of "dissimilar" samples are pushed further apart.

A cornerstone of many CL frameworks is data augmentation. By applying transformations (e.g., time warping, adding noise, channel dropout, frequency filtering for time-series; rotation, cropping for images) to an input sample (the "anchor"), multiple "views" are generated. These views are considered positive pairs, assumed to share core semantic information despite superficial differences. Samples originating from different source inputs are treated as negative pairs. The objective function, often based on noise-contrastive estimation principles like InfoNCE (Information Noise-Contrastive Estimation) [52], quantifies the similarity between positive pairs relative to negative pairs, guiding the model (typically an encoder network) to learn representations invariant to nuisance variability while retaining discriminative features.

A primary distinction among CL methodologies lies in their core learning objective, specifically concerning the use of negative samples. The traditional approach employs contrastive loss functions that explicitly require negative pairs. These objectives, such as the widely adopted InfoNCE loss [52], train the model to maximize the similarity between positive pairs (views of the same sample or corresponding cross-modal samples) while simultaneously minimizing their similarity to numerous negative pairs (samples from different instances or incorrect pairings). This process effectively aligns similar representations and separates dissimilar ones in the embedding space, often using cosine similarity as the metric for comparison. Alternatively, Non-Contrastive objectives learn representations without explicit negative samples, relying instead on mechanisms like self-distillation (e.g., predicting representations between augmented views using asymmetric networks) or optimizing statistical properties (e.g., reducing redundancy between embedding features) to prevent representation collapse while aligning positive pairs."

A critical challenge in contrastive learning is representation collapse, a scenario where the model produces identical or uninformative representations for all inputs. To prevent this, contrastive learning methods adopt either explicit or implicit strategies. Explicit contrastive methods incorporate negative samples into the loss function to push dissimilar representations apart, as seen in MoCo [28] (which uses memory queues) and SimCLR [29] (which relies on large batch sizes), these are typically referred to as instance discrimination methods. In contrast, implicit methods avoid the use of negatives. Some rely on architectural mechanisms—such as asymmetric networks and stop-gradient operations in BYOL [30], SimSiam [53], and DINO [54]—to break representational symmetry and prevent collapse; these are commonly categorized as self-distillation methods. Others embed collapse prevention directly into the objective function, as in Barlow Twins [55] and VICReg [56], which encourage statistical diversity and decorrelation in the learned features; these are typically referred to as correlation-based methods. For further details and theoretical foundations, readers are referred to [1].

**Instance discrimination**

Instance discrimination-based models treat training as a series of binary classification tasks, where a given data sample, referred to as the anchor, is compared against other samples categorized as either positive (similar/matching) or negative (dissimilar/non-matching).

In computer vision, instance discrimination methods like SimCLR [29] and MoCo [28] train models by applying random transformations to unlabeled images, creating augmented pairs. These images are then encoded into vector representations using an encoder and a momentum encoder, and a contrastive loss function is applied to minimize differences between positive pairs (augmentations of the same image) while maximizing differences between negative pairs (augmentations from different images). This approach enables models to learn robust representations that are invariant to trivial changes such as color, perspective, and occlusion, resulting in high generalization to downstream tasks.

### Non-contrastive collapse prevention

Despite the name, non-contrastive learning is closely related to contrastive learning—it simply removes the need for negative pairs. By relying only on positive pairs, these methods simplify training, requiring smaller batch sizes and avoiding the need for memory banks to store negative samples.

Non-contrastive collapse-prevention methods aim to train self-supervised models without using negative pairs, avoiding the challenges associated with contrastive learning while maintaining strong representation learning capabilities. These methods rely on alternative objectives to prevent representation collapse and ensure meaningful feature learning. Non-contrastive models like BYOL [30], SimSiam [53], DINO [54], Barlow Twins [55], and VICReg [56] have achieved results that are competitive with contrastive and supervised learning approaches, proving their effectiveness in representation learning.

Non-contrastive approaches can be categorized into two types: self-distillation methods and correlation-based methods [1]. Self-distillation methods replace contrastive loss with alternative training objectives, often using a student-teacher framework, where a student network learns to match the representations of a teacher network. Correlation-based methods prevent collapse by ensuring that the learned representations retain statistical diversity across dimensions. Instead of comparing different samples (as in contrastive learning), these methods optimize feature decorrelation and redundancy reduction within a single batch.

## B. Adapting contrastive learning to brain signals

The core principles of contrastive learning can be adapted to learn robust representations from complex brain signals by leveraging inherent similarities and differences within or across datasets. The fundamental idea is to maximize the similarity between augmented views of the same brain signal while minimizing the similarity between different signals. These adaptations generally involve three key components: data augmentation, positive (and optionally negative) pair selection, and the learning objective.

**Data augmentation:** This is the first step in contrastive learning and involves generating multiple views of the same EEG signal. Various augmentation techniques can be applied to create semantically consistent variations, including time-domain transformations (e.g., jittering, cropping), frequency-domain modifications (e.g., noise injection, band-pass filtering), and spatial transformations (e.g., channel shifting, amplitude scaling).

**Positive and negative pair selection:** The effectiveness of contrastive learning heavily depends on the appropriate construction of training pairs. Positive pairs are typically generated from different augmented views of the same EEG signal, while negative pairs are drawn from different signals or subjects. The quality of these selections directly impacts the model's ability to learn meaningful and discriminative neural representations.

**Learning objectives:** Following data augmentation and pair selection for brain signals, the learning objective defines the specific mathematical goal the model optimizes. Contrastive objectives, like the common InfoNCE loss [52], use both positive and negative pairs, explicitly training the model to pull positive representations closer while pushing negative representations apart, often using cosine similarity. Non-contrastive objectives, conversely, focus only on aligning positive pairs and rely on other mechanisms—such as self-distillation (e.g., student-teacher networks) or statistical regularization (e.g., maximizing variance, minimizing redundancy)—to prevent representation collapse without needing negative samples.

These adaptations of contrastive learning for brain data generally fall into two broad categories: unimodal approaches, which operate within a single brain modality (e.g., EEG or fMRI), and multimodal approaches, which integrate information either across multiple brain modalities (e.g., EEG and fMRI) or between brain signals and external non-neural modalities, such as text, images, or audio.

### Unimodal contrastive learning

The objective of unimodal contrastive learning is to learn representations that are invariant to irrelevant noise or superficial variations (e.g., session effects, minor electrode shifts) while remaining sensitive to the underlying neurophysiological patterns characteristic of that modality.
The core idea is to exploit the inherent structure and consistency within a specific type of brain recording. Neural signals like EEG often exhibit strong temporal continuity, where adjacent time segments share underlying patterns despite noise. Similarly, recordings from the same subject under similar conditions often show spatial consistency across channels or regions. CL leverages this by defining positive and negative pairs based on these relationships. Positive pairs are typically generated by applying different augmentations to the same original data segment (e.g., adding noise, time warping, frequency filtering, and channel dropout). For instance, EEGPT [40] utilizes perturbation-based CL, creating semantically consistent positive pairs through techniques like channel permutation and temporal jittering. Alternatively, positive pairs can be defined by temporal proximity; MBrain [57], for example, contrasts different time segments from the same multi-channel EEG/sEEG recording, enforcing temporal stability in the learned representations. Negative pairs usually consist of data segments derived from distinctly different time points, experimental trials, conditions, or even different subjects, forcing the model to learn discriminative features. Kostas et al., [58] adapt wav2vec 2.0's [59] contrastive predictive coding for EEG (BENDR), focusing on discriminating between different time segments of the same recording. Similarly, Li et al., [60] implement cross-subject contrastive learning with data augmentation, but specifically target motor imagery classification. Yang et al., [61] and Gijsen and Ritter, [62] implement collapse-prevention techniques without negative samples. Gijsen's approach is more comprehensive, incorporating both BYOL [30] and VICReg, [56] while Yang focuses on biosignal-specific adaptations.

By pulling representations of positive pairs closer while pushing negative pairs apart, the model learns to capture the stable, core patterns within the modality (e.g., characteristic oscillations, event-related potentials, spatial activation patterns) while discarding nuisance variability. This leads to more robust features for downstream tasks like sleep staging, seizure detection, or cognitive state classification based solely on that single brain signal type.

**Multimodal contrastive learning**

Multi-modal learning involves training models to align and integrate data from different modalities, such as text, images, audio, and video, enabling more robust and versatile representations. Contrastive learning has played a crucial role in multi-modal learning by enabling models to learn cross-modal relationships without requiring explicit supervision.

Contrastive methods learn to map relationships between modalities by jointly training separate encoders for each modality. For instance, Contrastive Language-Image Pretraining (CLIP) [63] trains an image encoder and a text encoder to predict which caption corresponds to which image using unlabeled (image, text) pairs collected from the internet. After pretraining, CLIP enables zero-shot learning, allowing NLP-based descriptions to reference visual concepts learned during training—even for images it has never seen before.

Beyond image-text learning, contrastive approaches have also been applied to other modality pairs, including video-text learning [64], which aligns video frames with textual descriptions; video-audio learning [65], which captures relationships between video content and background sounds; and speech-text learning [66], which maps spoken language to written text representations.

Multimodal contrastive learning encompasses CL approaches that learn by comparing or aligning brain signals with data from at least one other, different modality. The goal is typically to learn shared representations, find correspondences, or integrate complementary information across data streams. This can be further divided based on the nature of the other modalities involved.

**1. Cross-brain or physiological modality alignment**

This involves aligning signals from different types of brain recordings (e.g., EEG and fMRI) or aligning brain signals with other related physiological signals (e.g., EEG with ECG, EOG, EMG). Different brain recording techniques capture distinct but related aspects of neural function (e.g., EEG's high temporal resolution vs. fMRI's high spatial resolution). High-frequency EEG oscillations (like gamma waves) are known to correlate with fMRI BOLD signals during certain tasks [67]. Similarly, brain activity is intrinsically linked to other physiological processes. Aligning these signals via CL allows models to learn more comprehensive representations by integrating complementary information and finding shared underlying biological dynamics.

Directly aligning different brain modalities is an active area. For instance, Wei et al. [67] take a broad approach by implementing cross-domain and cross-modal self-supervised loss functions specifically designed to align representations derived from both fMRI and EEG data, aiming to capture shared functional information across these techniques. Extending beyond just brain signals, models like Brant-X [68] demonstrate aligning EEG with other physiological signals (ECG, EOG, EMG) using a two-level contrastive mechanism: local signal patches are projected into a shared space (patch-level alignment), and global temporal consistency is enforced across sequences (sequence-level alignment). These approaches facilitate robust representation learning across interconnected biological modalities.

**2. Brain-heterogeneous modality alignment**

This approach aligns brain signals (e.g., EEG, fMRI) with fundamentally different, external, non-neural modalities, most commonly images or text, but potentially also audio or video. This represents a significant shift from traditional supervised learning, which typically relies on predefined task labels (e.g., "motor imagery," "sad emotion"). By using CL to associate brain activity directly with the rich semantic content of images or text, models can learn representations grounded in real-world concepts. This allows for inferring underlying cognitive or perceptual states with much finer granularity (e.g., identifying what image a subject is seeing, or the conceptual content of language being processed, directly from brain signals) and enables potential zero-shot decoding capabilities.

Positive pairs consist of a brain recording and its corresponding external stimulus or description (e.g., fMRI scan + viewed image; EEG recording + presented text). Negative pairs involve mismatching the brain recording with incorrect external data. The contrastive loss drives the model to map corresponding pairs closer together in a shared or aligned embedding space.

Akbarinia, [71] and Song et al., [72] both focus on EEG-image alignment, but differ in their implementation. Akbarinia aligns EEG with pretrained CLIP features, while Song develops a self-supervised framework (NICE) specifically for image decoding. Kong et al. [76] aligned fMRI signals with visual stimuli features derived from the CLIP (Contrastive Language-Image Pre-training) model [63]. By optimizing fMRI embeddings to correspond to CLIP's image features using a contrastive objective, their model could infer visual scenes from brain activity. Similarly, NeuroLM [44] employs a cross-modal contrastive strategy in its initial stage to align EEG signals with textual representations, enhancing the model's ability to understand diverse cognitive states relevant to various Brain-Computer Interface tasks.

Wang et al., [49] combine contrastive learning with masked autoencoding, creating a multi-stream architecture for EEG-text alignment. Weng et al., [74] introduce a knowledge-driven approach combining cross-view contrastive methods with non-contrastive loss, incorporating neural domain knowledge.

Table 2. Summary of studies using SSL Contrastive methods applied to brain signals.

| Study | SSL Method | Key Features | Performance Metrics | Modality Support |
|---|---|---|---|---|
| Wei et al., 2025 [67] | Contrastive learning (Multi-modal Learning) | Cross-domain and cross-modal self-supervised loss | AUROC, Accuracy, Recall, Precision | EEG, fMRI |
| Ye et al., 2025 [69] | Contrastive learning (Instance Discrimination) | Maximizing mutual information for instance alignment | No mention found | EEG, Image |
| Zhou et al., 2024 [70] | Contrastive learning (Multi-modal Learning, Instance Discrimination) | BPE-level contrastive learning, Negative Contrastive Learning | BLEU scores, Accuracy | EEG, Text |
| Wang et al., 2024 [49] | Contrastive learning (Multi-modal Learning), Self-predictive learning (Masked Prediction) | Contrastive EEG-Text Masked Autoencoder | BLEU and ROUGE-1 scores | EEG, Text |
| Akbarinia, 2024 [71] | Contrastive learning (Multi-modal Learning) | Aligning EEG with pretrained CLIP features | Top-1 and Top-5 accuracies | EEG, Image |
| Gijsen and Ritter, 2024 [62] | Contrastive learning (Instance Discrimination, Non-contrastive Collapse-Prevention) | Multimodal contrastive InfoNCE loss, sub-unit alignment | Balanced accuracy, AUROC | EEG, Text |
| Li et al., 2024 [60] | Contrastive learning (Instance Discrimination) | Cross-subject contrastive learning | Accuracy | EEG |
| Song et al., 2024 [72] | Contrastive learning (Instance Discrimination) | Aligning image and EEG features | Top-1 and Top-5 accuracy | EEG, Image |
| Yang et al., 2024 [73] | Contrastive learning (Instance Discrimination) | Aligning brain signals with pretrained vision-language embeddings | Zero-shot classification accuracy, CLIP Score | EEG, fMRI, calcium imaging, spiking data |
| Yang et al., 2023 [61] | Contrastive learning (Non-contrastive Collapse-Prevention) | Biosignal Transformer with contrastive loss | Balanced Accuracy, AUC-PR, AUROC, Cohen's Kappa, Weighted F1 | EEG, ECG |
| Weng et al., 2023 [74] | Contrastive learning (Multi-modal Learning, Instance Discrimination, Non-contrastive Collapse-Prevention) | Knowledge-driven cross-view contrastive learning | Classification accuracy | EEG |
| Cai et al., 2023 [57] | Contrastive learning (Instance Discrimination) | Multi-channel Contrastive Predictive Coding | Precision, Recall, F1 score, F2 score, AUROC | EEG, SEEG |
| Banville et al., 2021 [75] | Contrastive: Instance Discrimination, SP: Innate Relationship (temporal context prediction) | Relative positioning, temporal shuffling, contrastive predictive coding | Balanced accuracy | EEG |
| Kostas et al., 2021 [58] | Contrastive learning (Instance Discrimination) | Adaptation of wav2vec 2.0 for EEG | Balanced Accuracy, AUROC | EEG |

In summary, multimodal CL involving brain signals pushes the boundaries of representation learning, either by integrating information from different biological recording techniques or by grounding neural activity in the rich semantic context of external data like images and text. These approaches are crucial for building more comprehensive and semantically aware Brain Foundation Models.

## 5. **Brain foundation models**

We adopt the definition introduced by Bommasani et al. (2021): "any model that is trained on broad data (generally using self-supervision at scale) that can be adapted to a wide range of downstream tasks" [17]. These models are typically large-scale, often containing hundreds of millions to billions of parameters, and are trained on massive, diverse datasets. Prominent examples include GPT-3/4 (GPT-3 [77] GPT-4 [78]) in NLP and CLIP [63] or Vision Transformers (ViT) [ref] in computer vision. A hallmark of foundation models is their versatility: a single pre-trained model can be fine-tuned or prompted to perform tasks it was not explicitly trained on.

In the context of this survey, a "brain foundation model" refers to a large neural model pre-trained on a wide range of brain recordings—spanning subjects, tasks, and datasets—using self-supervised learning. Such a model would serve as a general-purpose EEG feature extractor, enabling efficient adaptation to various downstream applications.

Recent work has begun developing foundation models for brain signals. These models are inspired by the success of foundation models in language and vision, and they aim to tackle the heterogeneity and label scarcity in brain signals and related neural data. Below we review several prominent foundation models for brain signals, including their architectures, data, pretraining strategies, and performance on downstream tasks.

Brant [38] is a large-scale model trained on invasive intracranial EEG (iEEG), specifically Stereo-EEG (sEEG), for tasks like seizure detection and neural signal forecasting. Brant-2 [79] and BrainWave [80] extend this to include both invasive iEEG and non-invasive scalp EEG. These models are characterized by high parameter counts and transfer learning capabilities across different recording modalities and tasks. BrainWave the lased version of them that pretrained on 13.8 TB of EEG+iEEG data (~40,000 hours from ~16,000 subjects), compared with 4TB and 1TB for Brant-2 and Brant, respectively. BrainWave introduces a hybrid architecture with an *embedding module* that segments each channel's time series into 1-second time-frequency patches, a multi-layer Transformer encoder that processes temporal sequences within each channel, and a *channel-attention* layer that integrates information across channels. This design allows BrainWave to handle varying recording lengths, sampling rates, and electrode layouts by adapting the patch embeddings for each signal's resolution. BrainWave is pretrained via self-supervised masked prediction that learns to reconstruct missing spectrogram patches. BrainWave has been evaluated across a wide range of clinical neurophysiology tasks, including cross-dataset neurological disorder diagnosis and seizure detection, which outperforming prior task-specific models. It demonstrates generalization ability, enabling zero-shot transfer to new hospitals or patient groups and few-shot learning with minimal data, highlighting the benefits of combining iEEG and EEG representations in one model.

BrainBERT [51] While the above models target EEG, BrainBERT (Wang *et al.*, 2023) focuses on *intracranial* neural recordings (iEEG/ECoG). Its architecture is based on a bidirectional Transformer encoder similar to BERT, adapted to neural time-series. To preprocess the data, BrainBERT uses high-resolution spectrograms of intracranial signals and applies a masking strategy akin to masked language modeling: segments of the time-frequency representation are randomly masked, and the model is trained to predict the missing content. This self-supervised pretraining was performed on a large corpus of unannotated iEEG recordings, allowing the model to learn a broad encoding of neural dynamics. In evaluations, BrainBERT generalized to new subjects with different electrode configurations and to entirely new cognitive tasks, indicating it learned fundamental features of intracranial signals rather than dataset-specific quirks. Overall, BrainBERT was a seminal step in creating foundation models for invasive recordings, showing that unsupervised Transformers can capture the latent structure of human brain activity.

BENDR (Kostas et al. 2021 [58]) is a self-supervised foundation model for EEG that draws inspiration from BERT [31] and wav2vec 2.0 [59]. The model processes raw multi-channel EEG signals through six convolutional layers—collectively termed the *BENDR Encoder*—to extract convolved feature representations. A subset of these features is then masked, and a Transformer encoder reconstructs the masked segments using the unmasked features as contextual information. The model is trained by minimizing a contrastive loss between the reconstructed and original convolved features. Building on BENDR, MAEEG (Chien et al. 2022 [81]) introduces architectural improvements that enhance performance. It adds a linear and a convolutional layer after the Transformer encoder to reconstruct the original EEG signal directly. Unlike BENDR's contrastive objective, MAEEG employs a reconstruction loss between the original EEG input and the reconstructed signal, leading to better signal-level fidelity.

Neuro-GPT (Cui et al., 2024 [47]) advances EEG representation learning by adopting the GPT framework [32]. It replaces the standard BENDR architecture with a decoder-only Transformer. Neuro-GPT consists of a convolutional module followed by a Transformer encoder—with an autoregressive GPT-style decoder. The model is pre-trained on a large-scale EEG dataset using a self-supervised objective of reconstructing masked time segments. In downstream evaluation, Neuro-GPT was fine-tuned on a motor imagery classification task with a small number of subjects. It outperformed models trained from scratch, demonstrating the advantage of using a pretrained foundation model in low-data EEG decoding scenarios.

GEFM (Graph-Enhanced EEG Foundation Model), by Wang et al. 2024 [43], improves over prior EEG foundation models by introducing a graph-enhanced architecture that captures spatial dependencies across EEG channels. Unlike models such as BENDR and Neuro-GPT, which primarily model temporal features, GEFM incorporates Graph Neural Networks to represent inter-channel relationships. This structural addition leads to more effective representation learning, with demonstrated performance gains across multiple downstream EEG classification tasks.

Another model, EEGFormer (Chen *et al.*, 2024 [42]), proposes an EEG foundation model pre-trained on a *compound* collection of many EEG datasets to learn *universal* EEG features. Experiments on various tasks (from seizure detection to waveform analysis) showed that EEGFormer achieved better performance, and its learned features could be used for anomaly detection and other applications without full retraining.

MOMENT, 385M parameter model, (Mononito Goswami *et al.*, 2024), is moving beyond domain-specific models by developing foundation models for general time-series analysis that can encompass neural data like EEG. MOMENT uses a Transformer architecture based on the T5 encoder-decoder and is pre-trained on a broad collection of time-series data drawn from many domains (a dataset dubbed the "Time-series Pile"). The pretraining task is masked time-series forecasting: the model learns to reconstruct or predict missing segments in a time series, which forces it to develop generalizable temporal representations. A key challenge addressed by MOMENT is the diversity of time-series characteristics (different sampling rates, lengths, variables) — the authors compile a large, diverse corpus of public datasets and devise training techniques to handle multi-dataset heterogeneity, enabling large-scale multi-domain pretraining for time-based data. Although not exclusive to brain signals, MOMENT has shown strong zero-shot and few-shot performance in time-series classification and anomaly detection. For example, in zero-shot evaluation, it achieved competitive or better accuracy than dedicated models on tasks like time-series classification, anomaly detection, signal imputation, and short-term forecasting.

Table 3. Brain foundation models.

| Name | Modality | Multi Tasks | Multi Modalities | Learning objective (Self-predictive or Contrastive) | Token Level (Region or Channel | Model Size Parameters (M) | Pretrained data size (TB) | Code |
|---|---|---|---|---|---|---|---|---|
| AnatCL [82], 2024 | EEG | ✓ | × | Contrastive | Region | 33 to 49 | 3,984 patients and 21,155 MRI images | https://github.com/EIDOSLAB/AnatCL |
| BENDR [58], 2021 | EEG | ✓ | × | Contrastive | Channel / Patch | 0.39 | 10,000 people and 1.5 TB | https://github.com/SPOClab-ca/BENDR |
| BRAINBERT [51], 2023 | EEG, audio | ✓ | × | - | Masked Modeling Channel | 43.18 | 4,551 electrode-hours | https://github.com/czlwang/BrainBERT |
| Brain-JEPA [37], 2024 | EEG | ✓ | × | JEPA (Self-predictive) | Region | 86 | 40,162 participants | https://github.com/Eric-LRL/Brain-JEPA |
| BrainLM [83], 2024 | EEG | ✓ | × | Masked / Autoregressive | Region | 13 to 650 | 77,298 (samples 6,700 hours) | https://github.com/vandijklab/BrainLM |
| BrainMAE [84], 2024 | EEG | ✓ | × | Masked Autoencoder (Self-predictive) | Region | - | 897 subjects | https://anonymous.4open.science/r/fMRI-State-F014 |
| BrainSegFounder [85], 2024 | EEG | × | ✓ | Masked / Contrastive | Region | 62 to 69 | 1,251 subjects | https://github.com/lab-smile/BrainSegFounder |
| BrainWave (Brant-2) [80], 2024 | iEEG, EEG | ✓ | ✓ | Masked Modeling / Contrastive? | Channel / Patch? | 1065M | 13.79 TB (16k sub 40,000 hours) | https://github.com/yzz673/Brant-2 |
| Brant [38], 2023 | iEEG (sEEG) | ✓ | × | Masked Modeling / Contrastive | Channel | 505.69M | 1.01 TB | https://github.com/yzz673/Brant |
| Brant-X [68], 2024 | EEG, EOG, ECG, EMG | ✓ | ✓ | Alignment / Contrastive | Channel / Patch | - | 4 | https://github.com/zjunet/Brant-X |
| CBraMod [39], 2025 | EEG | ✓ | × | - | Channel / Patch | - | 12 public datasets | https: //github.com/wjq-learning/CBraMod |
| CEReBrO [86], 2025 | EEG | ✓ | × | - | Channel / Patch | 3.6 to 85 M | 20,000 hours | - |
| EEGFormer [42], 2024 | EEG | ✓ | × | Self-predictive | Channel (electrode-level) | - | 1.7 | - |
| EEGPT [40], 2024 | EEG | ✓ | × | Autoregressive | Channel / Patch | 4.7M to 25M | - | https://github.com/BINE022/EEGPT |
| EEGPT-Yue [48], 2024 | EEG | ✓ | × | Autoregressive | Channel / Patch | 1.46Mto 1.09B | 37.5M samples, ~1B tokens | |
| FM-APP [87], 2024 | fMRI to sMRI | ✓ | ✓ | Knowledge Transfer / Contrastive | Region | - | 1200 individuals | https://github.com/ZhibinHe/FM-APP |
| FM-BIM [88], 2024 | EEG | ✓ | × | Masked / Contrastive | Region | - | 10 public datasets | - |
| FoME [89], 2024 | EEG | ✓ | × | - | Channel / Patch | 744.8M parameters | 1.7TB | - |
| LaBraM [10], 2024 | EEG | ✓ | × | Masked Modeling | Channel / Patch | 5.8M to 369M | 2,500 hrs (20 datasets) | https://github.com/935963004/LaBraM |
| LCM [90], 2025 | EEG | ✓ | × | Contrastive | Channel / Patch? | 33.9M | 109 subjects | - |
| MBrain [57], 2023 | EEG | × | ✓ | Contrastive / Masked | Channel | - | 550GB | - |
| MEET [91], 2024 | EEG | ✓ | × | Masked / Contrastive | Channel / Patch | 30 M to 215M | - | - |
| MeTSK [92], 2023 | EEG | ✓ | × | Contrastive / Masked | Region | - | 1,096 subjects | - |
| Neuro-GPT [47], 2024 | EEG | × | × | Autoregressive | Channel / Patch | 79.53 | 5,656 hours | https://github.com/wenhui0206/NeuroGPT |
| NeuroLM [44], 2025 | EEG, text | ✓ | ✓ | Masked / Autoregressive | Channel / Patch + Text Token | 254M to 1,696M | 25,000-hours | https://github.com/935963004/NeuroLM |
| NeuroVNN [93], 2024 | EEG | × | × | - | Region | - | 2,147 healthy individuals | |
| TF-C [94], 2022 | EEG, ECG, EMG bearings, human activity, hand gestures | ✓ | ✓ | Contrastive | Time Patch / Segment | - | 8 datasets | https://github.com/mims-harvard/TFC-pretraining |
| TGBD [76], 2025 | EEG, images | × | × | - | Region | - | 177 subjects | https://github.com/Xiangtaokong/TGBD |

*iEEG: Intracranial electroencephalography (intracranial EEG) includes: sEEG: Stereo-electroencephalography (Stereo-EEG) and ECoG: Electrocorticography.*

LaBraM (Large Brain Model) [10], introduced by Jiang *et al.* (2024), focuses on EEG signals and introduces a unified model trained across multiple heterogeneous EEG datasets. It employs a two-stage training process involving a neural tokenizer and a masked modeling objective. LaBraM introduces a *neural tokenizer*: a vector-quantization module is trained to convert continuous EEG patches into discrete codebook tokens by predicting their power spectrum, creating a vocabulary of neural "words". The main Transformer is then pretrained to predict masked tokens from context (analogous to BERT) using a massive EEG corpus (~2,500 hours from ~20 diverse EEG datasets). This two-stage self-supervised learning (tokenizer + masked modeling) yields semantically rich, dataset-agnostic EEG representations. LaBraM was evaluated on heterogeneous downstream tasks – including abnormal EEG detection (e.g. identifying pathological vs. normal EEGs), event-type classification, emotion recognition, and even gait pattern

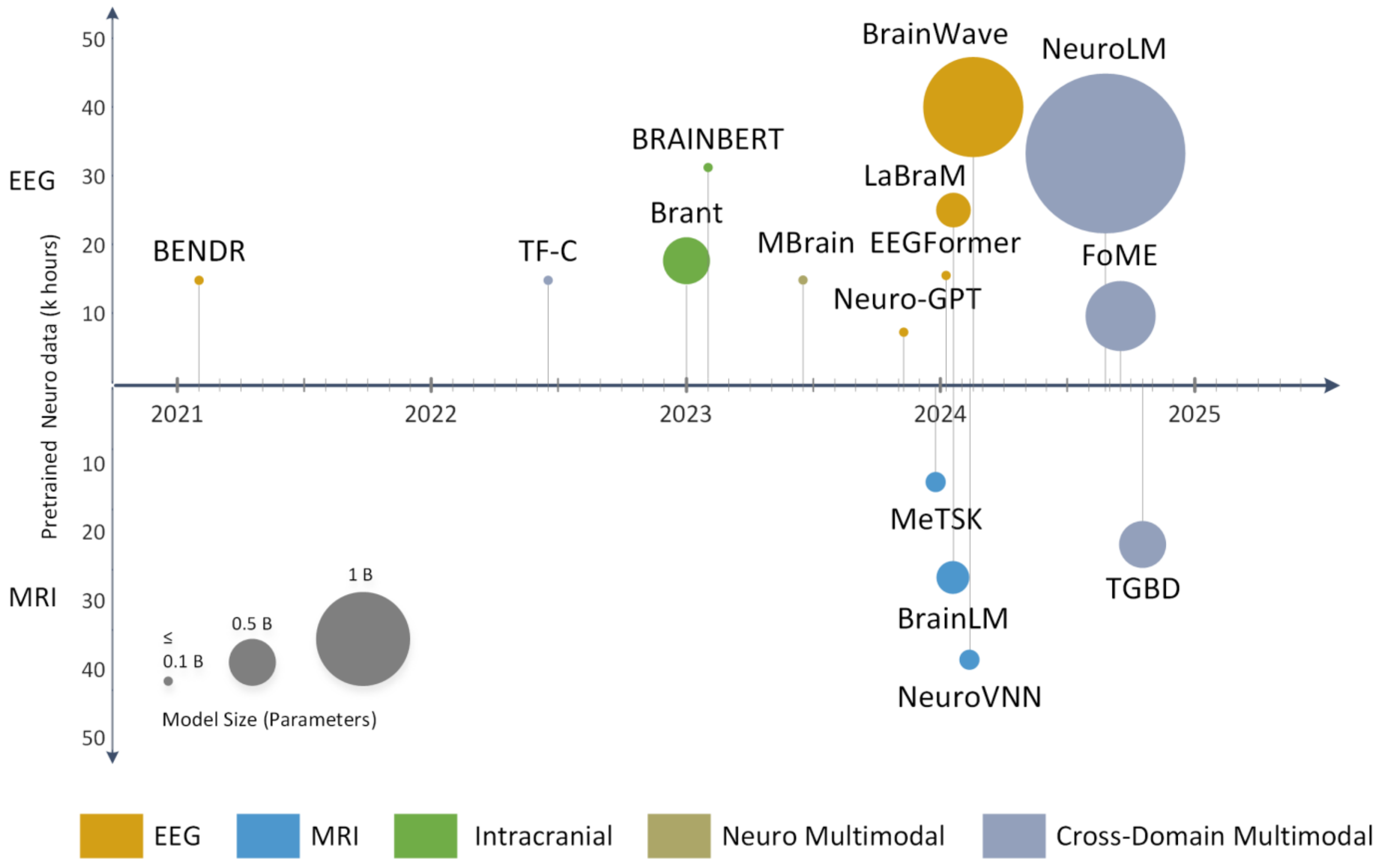

Figure 10. Timeline and characteristics of recent foundation models in brain-related AI research. The x-axis represents the publication year, while the y-axis shows the approximate amount of neuro data (in thousands of hours) used for pretraining. Circle size corresponds to model size (parameter count), with reference sizes shown in the bottom left. Colors indicate the modality: EEG (yellow), MRI (blue), Intracranial (green), Neuro Multimodal (olive), and Cross-Domain Multimodal (gray-blue). Notable models such as BENDR, BrainWave, NeuroLM, and LaBraM are displayed to illustrate the growing trend and diversity in neuro foundation model development across data modalities and scales.

prediction, achieving significant performance improvements through its learned EEG representations.

Recently, researchers have begun merging foundation EEG models with large language models to enable multi-task learning. NeuroLM (Jiang *et al.*, 2025) [44] is an ambitious example that treats EEG as a "foreign language" to be processed by an LLM. Compared to earlier models like LaBraM [10], which use task-specific fine-tuning, NeuroLM improves generality and scalability by using EEG-Text alignment and LLM Integration. NeuroLM first encodes raw EEG into discrete tokens via a learned VQ (vector-quantized) tokenizer (akin to words for EEG) and then feeds these tokens into a Transformer language model that has been adapted to handle EEG sequences in an autoregressive manner. Through multi-task instruction tuning, NeuroLM can perform a variety of EEG analysis tasks (such as decoding stimuli, classifying mental states, etc.) within one unified model. The largest version, NeuroLM-XL, has 1.7 billion parameters and was pre-trained on ~25,000 hours of EEG recordings – the largest EEG corpus to date. Evaluated on six different EEG datasets, NeuroLM showed promising multi-task performance. This approach represents a convergence of neuro and language domains, suggesting a new frontier where foundation models not only learn from neural signals in isolation but also align them with semantic frameworks used in NLP.

## Evaluation of SSL-FMs

The most common way to assess an SSL-pretrained model is to use it for a downstream EEG decoding task, such as motor imagery (MI) classification. Several evaluation strategies are typically employed:

- Linear Evaluation (known as liner-probing evaluation): The pre-trained backbone is frozen, and only a linear classifier is trained on top of the learned representations. This method tests whether the learned representations are linearly separable, providing a fast and efficient way to assess feature quality.
- Nonlinear Evaluation: A more complex nonlinear classifier (e.g., MLP, SVM) is added.
- Fine-Tuning Evaluation: The entire model, including the SSL-pretrained backbone, is fine-tuned on labeled EEG data.

Currently, most brain foundation models depend extensively on task-specific fine-tuning, which seems somewhat contrary to the ultimate goal of SSL—learning universal representations that generalize effectively across diverse downstream tasks with minimal labeled data. While SSL provides a powerful pre-training framework, the near-universal need for considerable fine-tuning suggests that the pre-trained representations, while beneficial, may not yet fully capture the kind of versatile, zero-shot (KNN or liner-probing evaluation) or few-shot capabilities demonstrated by LLMs like GPT in the NLP domain. Even NeuroLM [44], which represents a pioneering effort to mitigate fine-tuning through multi-task instruction tuning, shows improved adaptability but reportedly still lags behind task-specific fine-tuned models in performance benchmarks. This highlights that current SSL strategies for foundation models have yet to bridge the gap toward truly task-agnostic, general-purpose neural representations with zero-shot capabilities.

Compared with the transformative impact of the foundation models in NLP and computer vision, their impact on EEG-based brain signal analysis remains limited. These models are typically designed for structured, discrete data with relatively stable patterns (e.g., text or images). EEG data, however, is irregular, dynamic, and continuous, making it poorly suited for standard attention-based architectures without significant adaptation [95], [96]. In addition, FMs generally rely on large-scale, diverse datasets to achieve robust generalization. However, EEG datasets are typically limited in size, imbalanced, and expensive to collect [97], [98]. Recent EEG foundation models, including BENDR (86.7%), Brain-JEPA (81.52%), and BrainSegFounder (91.15%) demonstrate potential in terms of accuracy, but they still fall short of their NLP and CV counterparts, particularly when tested on a variety of downstream tasks. Pretraining foundation models on EEG therefore necessitates aggregating heterogeneous data from multiple sources—introducing challenges related to variability in experimental setups, recording conditions, and subject populations, all of which complicate training and limit scalability.

## 6. Multimodal integration

Multimodal integration is an evolving area in brain signal decoding, aimed at leveraging complementary information from diverse data sources to improve performance, generalizability, and robustness. Key directions include the integration of brain imaging modalities, broader physiological biosignals (e.g., ECG, EMG), and external modalities such as language, visual, and auditory data.

### 6.1 Brain and biosignal integration

Several studies have investigated the integration of diverse brain and physiological signals to enhance decoding performance and representation learning. These include EEG and fMRI integration [67], [99], [100], multiple neural signal integration involving modalities such as EEG, MEG, and fMRI [101] [73], and biosignal integration combining EEG with other physiological signals like ECG [61] and EMG [50]. The following subsections provide an overview of representative works across these categories.

**EEG and fMRI integration**: These studies aim to combine the high temporal resolution of EEG with the high spatial resolution of fMRI to gain more comprehensive insights into brain function. This integration serves two main objectives: enhancing SSL representations [67] and enabling EEG-to-fMRI signal generation [99], [100]. Wei et al. [67] introduce a Multi-modal Cross-domain Self-supervised Pre-training Model (MCSP) that fuses EEG and fMRI data across spatial, temporal, and frequency domains, improving the quality of learned representations through joint contrastive learning. In the context of cross-modal generation, Li et al. [99] present a method for EEG-to-fMRI synthesis by utilizing LaBraM [10] pretrained model, while [100] propose a Condition-Aligned Temporal Diffusion (CATD) framework that aligns EEG and fMRI distributions to generate fMRI signals conditioned on EEG inputs. These approaches highlight the potential of EEG-fMRI integration to advance both cross-modal representation learning and neural signal generation.

**Multiple neural signal integration:** These approaches aim to develop unified representations across diverse neuroimaging modalities, potentially enabling more comprehensive brain state decoding. Ferrante et al. [101] focus on aligning representations from EEG, MEG, and fMRI, while Yang et al. [73] propose a framework for learning joint representations across EEG, fMRI, spiking data, and calcium imaging.

**Biosignal integration**: Broader research efforts have focused on integrating multiple biosignals—such as EEG, ECG, and EMG—to develop more generalizable models capable of learning across diverse physiological data. These approaches aim to enhance model robustness and adaptability in real-world biomedical applications. Yang et al. [61] propose BIOT, a cross-data biosignal learning framework designed to handle heterogeneous signals including EEG, ECG, and others. Similarly, Wu et al. [50] explore SSL techniques that simultaneously leverage EEG and EMG data, demonstrating the potential of SSL for multi-modal biosignal representation learning.

### 6.2 Neural-text integration

An increasing number of studies are exploring the integration of brain signals with linguistic information to enable more semantically meaningful EEG decoding and interpretation. This line of research can be categorized into three areas:

**EEG-to-text decoding**: Studies such as Amrani et al. [102] and Wang et al. [49] have developed frameworks to translate EEG signals directly into textual representations, using models like contrastive masked autoencoders to align EEG patterns with language data.

**Language model integration**: Other efforts incorporate LLMs to improve the interpretation of EEG signals. For example, NeuroLM [44] introduces a unified multi-task foundation model that encodes EEG into discrete tokens via a vector-quantized tokenizer and aligns them with text embeddings using adversarial training—without the need for paired EEG-text data. Gijsen and Ritter [62] further explore EEG-language modeling using clinical EEG reports.

**Cross-modal alignment**: Some approaches focus on aligning EEG features with textual embeddings at various levels. Zhou et al. [70] introduce a subword-level (BPE) alignment method for EEG-language fusion, while Zhou et al. [103] explore more generalized cross-modal alignment strategies.

These studies aim to integrate EEG with natural language representations, paving the way for more accurate, interpretable, and semantically enriched brain signal decoding.

## 6.3 Neural-visual integration

The integration of visual information with brain signals is an active area of research, focusing on understanding the correspondence between neural activity and visual stimuli. This work can be grouped into the following categories:

**EEG-image alignment**: Studies such as Song et al. [72], Palazzo et al. [104], and Ye et al. [69] investigate methods for associating EEG signals with visual stimuli. Song et al. [72] present a self-supervised framework for EEG-based image decoding, while Palazzo et al. [104] explore multimodal learning of neural activity and visual features.

**Cross-modal visual decoding**: Akbarinia [71] and Du et al. [105] explore the use of EEG to classify or reconstruct visual information, including contrastive learning and signal representation techniques [106], [107]. Akbarinia [71] applies contrastive learning to align EEG features with pre-trained CLIP features. Du et al. [105] propose A multimodal neural decoding method that combines brain, visual, and linguistic features.

## 6.4 Neural-audio integration

These studies explore the relationship between brain activity and auditory processing, with potential applications in speech decoding and auditory attention detection. Li et al. [108] develop a multi-stage strategy for semantic brain signal decoding, incorporating audio-text alignment.

**Multimodal Foundation Models:** These studies aim to develop foundation models that can handle multiple modalities, potentially enabling more flexible and powerful decoding approaches. Lei et al. [109] propose ViT-LENS for omni-modal representations, including EEG. Fang et al. [45] work on promoting cross-modal representations in multimodal foundation models for physiological signals.

**Key techniques for multi-modal integration:**

Many studies use contrastive learning to align representations from different modalities. For example, Akbarinia [71] and Song et al. [72] use this approach for EEG-image alignment. Some studies employ cross-modal autoencoder frameworks to learn joint representations of multiple modalities, such as the approach used by Du et al. [105]. Many multi-modal approaches leverage transformer architectures for their ability to handle diverse input types [109]. For instance, Yang et al. [73] use a Vision Transformer (ViT) as the backbone for their multimodal model. Techniques like vector quantization are used in some studies to create discrete representations that can be aligned across modalities, as seen in Jiang et al. [44].

**Challenges and limitations:**

Multimodal integration with EEG data introduces several technical and methodological challenges that must be carefully addressed to build effective and generalizable models. A fundamental issue is the mismatch in temporal and spatial resolution between EEG and other modalities such as fMRI, MEG, or natural language [67], [99]. EEG provides high temporal but low spatial resolution, whereas other modalities may offer complementary characteristics, making temporal and spatial alignment non-trivial. Additionally, determining how to optimally fuse information across modalities—whether through early, late, or hybrid fusion strategies—remains an open research question, with no one-size-fits-all solution [101], [105], [109].

Scalability is another concern, as multimodal models typically require significantly more computational resources and large, diverse datasets, which are not always available, particularly in medical and neuroscience applications [44], [48]. The increased complexity of these models also raises concerns about interpretability. As multimodal architectures grow in depth and scope, understanding their internal decision processes becomes more difficult, which is problematic in clinical or high-stakes applications [42]. Finally, each modality introduces its own types of noise and artifacts—for example, EEG is susceptible to muscle movement and electrical interference, while imaging or language data may carry contextual or semantic noise—making robust integration an additional layer of complexity [57], [61]. These factors highlight the need for continued exploration of principled, efficient, and interpretable approaches to EEG-based multimodal learning.

# 7. Adapting foundation models for new tasks

Once the foundation model has been pretrained using SSL techniques on an upstream or pretext task, the next steps involve evaluating the quality of the learned representations and adapting the model for specific, practical applications (downstream tasks). This evaluation and adaptation process is fundamental to assessing the foundation model's effectiveness and generalizability. The utility of these models is measured by their performance on tasks such as motor imagery (MI) classification, cognitive state recognition (e.g., emotion detection), and clinical applications like seizure detection or sleep stage classification. This section details the common protocols for evaluating SSL-pretrained representations and the standard workflows for adapting these powerful models to targeted downstream neuroscience and clinical challenges.

## 7.1 Assessing representation quality: evaluation protocols

Before extensive task-specific adaptation, it's essential to assess the quality of the features learned during SSL pretraining. Evaluation is typically performed using one or more of three main approaches, ranked by increasing complexity: k-nearest neighbors (KNN), linear evaluation, and full fine-tuning [1].

Table 4. EEG Datasets for SSL Tasks.

| Name (URL), Pub. Year | Dataset type | # channels | # subjects | Tasks |
|---|---|---|---|---|
| EEG-VOA[1] [110], 2017 | EEG, Images | 128 | 6 | Passive (Visual: Images) |
| ETCAS[2] [111], 2023 | EEG, Speech | 24 | 50 | Passive (Auditory: Speech) |
| MusicAffect[2] [112], 2022 | EEG, Music | 31 | 21 | Passive (Auditory: Music) |
| ASD dataset [113], 2023 | EEG | 20 to 129 | 4899 | AS-Disorder |
| parrKULee [114], 2023 | EEG | 64 | 85 | Speech decoding |
| CUHZ [115], 2022 | EEG | 22 | 25 | Seizure detection |
| Visual object[3] [116], 2022 | EEG | 64 | 10 | Image decoding |
| MPI-LEMON[4] [117], 2019 | EEG, MRI, ECG, etc. | 62 | 216 | Non Resting states |
| KU-MI [118], 2019 | EEG, EMG | 62 | 52 | Motor imagery |
| MPED [119], 2019 | EEG, ECG, ESR, RSP | 62 | 23 | Emotion recognition |
| Physionet Challenge [120], 2018 | EEG, EMG, EOG, etc. | 6 | 1983 | Sleep classification |
| AMIGOS , 2018 | EEG, ECG, GRS | 14 | 40 | Emotion recognition |
| SEED-IV [121], 2018 | EEG | 62 | 15 | Emotion recognition |
| TUH abnormal[5] [122], 2017 | EEG | 27 to 36 | 2329 | Abnormal detection |
| DREAMER [123], 2017 | EEG, ECG | 14 | 23 | Emotion recognition |
| SEED [124][124], 2015 | EEG | 62 | 15 | Emotion recognition |
| Mayo-UPenn Seizure Dataset [125], 2015 | EEG, Dog signal | 16 | 4 | Seizure detection |
| MASS [126], 2014 | EEG, EOG, EMG, ECG | 20 | 62 | Sleep classification |
| BCIC [127], 2012 | EEG, EOG | 22 | 9 | Motor imagery |
| DEAP [35], 2011 | EEG, Video, EOG, EMG | 32 | 32 | Emotion recognition |
| MAHNOB-HCI [128], 2011 | EEG, Multiple signals | 32 | 27 | Emotion recognition |
| CHB-MIT [129], 2019 | EEG | 24-26 | 24 | Seizure detection |
| MMI [130], 2004 | EEG | 64 | 105 | Motor imagery |
| Epilepsy Dataset [131], 2001 | EEG | 19 | 500 | Seizure detection |

1 https://github.com/perceivelab/eeg_visual_classification?tab=readme-ov-file
2 https://doi.org/10.6084/m9.figshare.12326519.v1
3 https://osf.io/3jk45/
4 https://ftp.gwdg.de/pub/misc/MPI-Leipzig_Mind-Brain-Body-LEMON/
5 https://isip.piconepress.com/projects/nedc/html/tuh_eeg/

**K-Nearest neighbors (KNN) evaluation:** This simple approach involves feeding downstream task data through the frozen pretrained backbone to extract features. A KNN classifier is then trained on these features. High KNN accuracy suggests that the SSL pretraining has effectively clustered semantically similar inputs in the feature space, even without linear separability. It provides a basic, low-complexity check on feature organization.

**Linear evaluation (linear probing):** This is arguably the most common protocol for evaluating SSL representations. The pretrained backbone model is frozen (its weights are not updated), and only a simple linear classifier (typically a single fully connected layer) trained on top of the extracted features is optimized for the downstream task using labeled data. Strong performance indicates that the SSL pretraining has learned linearly separable features relevant to the task. It isolates the quality of the features themselves without complex adaptation confounding the results.

**Full fine-tuning:** This involves unfreezing some or all layers of the pre-trained backbone and training the entire model (backbone + task-specific head) end-to-end on the downstream task's labeled data, often using a lower learning rate for the backbone layers. While fine-tuning usually yields the highest performance, it also assesses the model's adaptability and the combined effect of pretraining and task-specific adaptation, rather than solely the inherent quality of the initially learned representations.

**Evaluation metrics:** Performance assessment relies on task-appropriate metrics. Classification tasks commonly use accuracy, balanced accuracy (crucial for imbalanced data like seizure detection), Area Under the Receiver Operating Characteristic curve (AUROC), F1-score, and Cohen's Kappa (for sleep staging). Generative tasks like brain-to-text use metrics from NLP, such as BLEU and ROUGE scores. Visual decoding often uses classification accuracy (Top-1, Top-5) or image reconstruction metrics. The diversity of metrics across studies can sometimes make direct model comparison challenging.

## 7.2 Standard adaptation workflow for downstream tasks

Adapting a pretrained foundation model for a specific downstream application generally follows the following steps:

1. **Remove Pretraining-Specific Layers:** SSL methods often utilize auxiliary layers during pretraining, such as projection heads (MLPs used in contrastive learning like SimCLR/MoCo) or decoder heads (used in masked prediction like MAE). These layers are optimized for the SSL objective (e.g., contrastive separation, reconstruction) and are generally not beneficial for downstream discriminative or generative tasks. Therefore, they are typically discarded before adaptation.
2. **Add a Task-Specific Head:** New layers tailored to the downstream task are added on top of the pre-trained backbone (the core feature extractor). For classification tasks, a fully connected (FC) layer or a multi-layer perceptron (MLP) with an appropriate output size (number of classes) and activation function (e.g., softmax) is common. For regression tasks: an FC layer with a linear activation might be used. In the case of generation/decoding, more complex decoder architectures might be added for tasks like brain-to-text translation.
3. **Select a Training Strategy** (Fine-tuning vs. Frozen Backbone): A choice is made on how to train the combined model (backbone + new head) on the downstream labeled data.
   <u>In frozen backbone mode</u>, the weights of the pre-trained backbone remain frozen. Only the newly added task-specific head layers are trained. This is computationally efficient, requires less labeled data, and reduces the risk of overfitting or catastrophic forgetting of the pre-trained knowledge. It's effective if the pre-trained features are already highly suitable for the task.
   <u>Full Fine-tuning Mode</u>: All or part of the pre-trained backbone is unfrozen (often starting with the top layers), and its weights are updated along with the new head, typically using lower learning rates for the backbone to preserve pretrained knowledge. This allows the backbone to adapt its features more closely to the nuances of the downstream task, potentially leading to higher performance but requiring more labeled data and careful hyperparameter tuning.

## 7.3 Common downstream tasks

The pretrained foundation models are tested across the following tasks:

**Clinical applications:**

**Seizure detection/classification:** Identifying epileptic seizure events from EEG recordings. Examples include evaluations by Shi et al. [89] (FoME), Yang et al. [61] (BIOT), and Cai et al. [57] (multi-channel CPC).

**Sleep stage analysis:** Classifying segments of EEG data into different sleep stages (e.g., Wake, REM, N1, N2, N3). Examples include work by Ogg and Coon [132], Banville et al. [75], and Shi et al. [89].

**Abnormal EEG detection:** Identifying deviations from normal background EEG activity, often related to various neurological conditions [10], [42], [48].

**Cognitive state decoding:**

**Emotion recognition:** Classifying emotional states based on EEG patterns elicited by stimuli or internal states. Examples include Jiang et al. [10] (LaBraM), Yue et al. [48] (EEGPT), and Cai et al. [57].

**Motor imagery classification:** Decoding imagined movements (e.g., left vs. right hand) from EEG, crucial for Brain-Computer Interfaces (BCIs). Examples include Cui et al. [47] (Neuro-GPT), Li et al. [60], and Yue et al. [48].

**Event-related potential (ERP) detection:** Identifying specific brain responses time-locked to sensory, cognitive, or motor events.

**Cross-modal decoding (generative & interpretive tasks):** Exploring the potential of SSL-pretrained models beyond classification to generate outputs in different modalities based on brain activity.

**Brain-to-Text translation:** Generating textual descriptions corresponding to cognitive states or perceived stimuli directly from EEG/MEG signals. For example, Meta AI researchers [133] explored decoding speech perception from non-invasive brain recordings (EEG/MEG) by correlating brain activity with corresponding speech segments using a CLIP-like framework. In this setup, neural signals were recorded while subjects listened to speech, enabling alignment between the auditory stimuli and neural responses. Other examples include the work of Wang et al. [49], Zhou et al. [70], and Amrani et al. [102], who have demonstrated various approaches for aligning brain signals with textual representations to enable EEG-to-text decoding.

**Visual decoding:** Reconstructing or classifying visual stimuli perceived by a subject based on their brain signals. Examples include Song et al. [72], Akbarinia [71], and Jing et al. [134] (LLMs for EEG visual decoding).

The effectiveness of these models varies across different downstream tasks, with larger foundation models generally showing better transfer learning capabilities but requiring more computational resources. Multi-modal approaches tend to offer more flexible applications across different types of tasks but face additional challenges in integration and alignment.

### 7.4 Foundation models adaptation beyond fine-tuning

While fine-tuning is the most common adaptation method, pretrained foundation models can be leveraged in other ways, sometimes categorized under the umbrella of "Learning From Models" (LFM) [134]. This includes using the frozen backbone purely as a feature extractor (as discussed in linear probing and feature extraction mode), potentially employing knowledge distillation techniques, or exploring prompting strategies if the model architecture allows. These alternative approaches offer different trade-offs in terms of performance, computational cost, and data requirements. For a detailed exploration of LFM strategies, readers are referred to [134].

### 7.5 Datasets

In the realm of SSL EEG data, various datasets serve as foundational resources for model training and evaluation. In Table 5, we explore several EEG datasets that have been utilized in recent studies. These datasets can be categorized into: Passive (Stimuli-Evoked) Neuroimaging Datasets and Active (Task-Based) Neuroimaging Datasets.

**Passive (stimuli-evoked) neuroimaging datasets:** Datasets where participants are passively exposed to external stimuli without performing any active task (no overt response or intention required). They capture brain responses to visual and auditory stimuli—such as images, videos, sounds, speech, and music—including, for example, EEG recordings while watching movies or listening to music. These datasets are commonly used in conditional multimodal synthesis, where a generative model is trained to produce data in one or more modalities (e.g., image, audio, or text) conditioned on input from another (e.g., brain signals) [18]. In this context, the brain response serves as a condition to reconstruct or generate the original stimuli, aiming to infer what the subject was seeing or hearing based solely on their neural activity.

**Active (task-based) neuroimaging datasets:** These datasets involve scenarios where participants are actively engaged in performing specific mental or physical tasks, often without the presence of external sensory stimuli. Tasks may include motor imagery, mental arithmetic, or language production. Active neuroimaging datasets are frequently used to decode internal cognitive states or intentions from brain signals, such as MI classification.

## 8. Discussion and future directions

### 8.1 Self-supervised learning beyond EEG?

Self-supervised learning (SSL) methods have yielded promising results on EEG, but performance often remains lower than on related modalities like MEG. Studies consistently report that MEG outperforms EEG in similar tasks, largely due to its higher signal-to-noise ratio and superior sensor coverage. For instance, a contrastive learning decoder trained to identify speech segments from brain activity achieved up to 71% accuracy with MEG, compared to just 18–26% with EEG [133]. In another study focused on decoding sentence production while participants typed on a QWERTY keyboard [135], the proposed Brain2Qwerty model achieved a 32% character error rate using MEG, versus approximately 67% with EEG—again highlighting the performance gap. These findings emphasize that EEG's lower fidelity presents a significant challenge for foundation models, often demanding more data or advanced methods to approach MEG-level results [133]. However, existing MEG systems are not portable. This limitation may be addressed by emerging MEG technologies based on optically pumped magnetometers (OPMs) [135], [136].

Notably, the performance hierarchy between modalities can depend on the task. While MEG generally excels in decoding tasks such as continuous speech [133] or text typing [135], some evidence suggests EEG can match or even surpass MEG for certain visual decoding problems. For instance, an image-reconstruction study [137] found EEG-based models performed on par with MEG in classifying viewed images, contrary to the typical speech decoding trends. Overall, however, MEG's advantages make it the more performant non-invasive modality in most SSL settings, implying that insights gained from MEG experiments set an upper bound for what purely EEG-based self-supervised models can achieve.

A related modality, intracranial EEG (iEEG), differs significantly from scalp EEG in signal quality and spatial resolution, which has implications for SSL. While EEG is non-invasive and widely accessible, it suffers from low spatial resolution and a poor signal-to-noise ratio due to attenuation by the skull and scalp. In contrast, iEEG—being invasive—offers significantly higher signal fidelity and spatial resolution by recording directly from cortical surfaces or within the brain tissue [80]. These differences allow iEEG to capture more precise neural patterns, leading to superior performance in clinical diagnostic tasks. For instance, the BrainWave foundation model jointly pre-trained on both modalities showed that iEEG data contributed more significantly to performance improvements than EEG alone, particularly in cross-domain and few-shot classification tasks. Notably, the inclusion of iEEG in joint pretraining led to an average improvement of 7.44% in AUROC over an EEG-only model, emphasizing the strength of iEEG in learning generalizable neural representations [80]. However, the invasiveness of iEEG limits its applicability to clinical contexts, underscoring a trade-off between performance and practicality.

EEG and fMRI are two widely used neuroimaging modalities that offer complementary insights into brain function. EEG captures electrical activity directly from the scalp with millisecond-level temporal resolution, making it highly effective for tracking rapid neural dynamics.

However, it suffers from low spatial resolution due to signal diffusion across the scalp and skull. In contrast, fMRI measures slow fluctuations in blood-oxygen-level-dependent (BOLD) signals, providing high spatial resolution suitable for mapping brain-wide functional networks, but with limited temporal precision. This contrast—EEG's strength in time and fMRI's strength in space—motivates the integration of both modalities for a more comprehensive understanding of brain disorders. The Study by Wei et al. [67] proposed leverages this synergy through cross-domain and cross-modal self-supervised learning, showing that the fusion of EEG and fMRI can enrich feature representations and improve classification performance across multiple psychiatric conditions.

## 8.2 EEG-to-Text models: are they really working?

The growing interest in EEG-to-Text models has led to a wave of studies that aim to decode natural language directly from brain activity, particularly using pre-trained language models (e.g., BART [], T5 [], PEGASUS []). These models often follow a sequence-to-sequence (seq2seq) architecture, where EEG-derived embeddings—typically collected during reading tasks—are mapped to corresponding textual sequences. However, the reliability of these methods has come under scrutiny.

The recent study by Jo et al. (2024) [138], raises critical concerns about whether these models are genuinely learning from EEG signals or merely memorizing label patterns. A key methodological issue identified is the use of teacher forcing during both training and evaluation, which involves feeding the ground-truth tokens to the model at each step of text generation [139]. While this improves training stability, using it during evaluation artificially boosts performance. When the authors removed teacher forcing at inference time and instead used standard generation, performance dropped significantly—suggesting that earlier high-performance metrics were likely inflated.

Additionally, the authors conducted noise ablation studies, showing that models trained and evaluated on random noise performed similarly to those trained and tested on actual EEG data. This suggests that the models were not learning meaningful representations from EEG inputs, but rather leveraging statistical regularities from the language model itself.

Several recent studies have adopted similar EEG-to-Text decoding protocols and are potentially affected by these issues, including the papers refereed by [140]–[144]. These models also rely on language generation from EEG embeddings and employ similar evaluation schemes involving seq2seq learning. Jo et al.'s findings raise concerns about whether these models genuinely decode semantic content from EEG or simply leverage linguistic priors.

The results highlight the lack of input sensitivity in current EEG-to-Text pipelines, suggesting a need for more rigorous, controlled, and reproducible protocols, including comparison against noise-based baselines, Evaluation with and without teacher forcing, cross-condition tests (e.g., training on EEG, testing on noise).

## 8.3 Efficacy of self-supervised learning for brain foundation models

Self-supervised learning has undeniably become the predominant pre-training paradigm for developing large brain models. However, despite its successes in domains like natural language processing and computer vision, the direct translation and ultimate effectiveness of SSL methodologies for the complexities of neural data warrant critical examination. While SSL thrives on the structured information and rich semantic content inherent in language (e.g., syntax, co-occurrence patterns exploited by models like BERT [31]) and images (e.g., visual concepts aligned with text by CLIP [63]), brain signals present a starkly different landscape. Neural data, encompassing modalities like EEG, fMRI, and iEEG, is notoriously characterized by high levels of noise (both physiological and measurement-related), substantial trial-to-trial and inter-individual variability, and often lacks the clearly defined, hierarchical structure found in text or visual scenes. These fundamental differences pose unique and significant challenges to the standard application of SSL techniques.

<u>1. Challenges arising from neural data characteristics:</u>

A primary hurdle stems from the inherent instability and variability of neural activity. Even under identical experimental conditions, brain patterns can differ considerably across individuals and even within the same individual over time. This makes establishing a consistent, generalizable representation space—a prerequisite for effective large-scale learning—remarkably difficult. Standard SSL techniques often implicitly assume a degree of stability or semantic consistency that may not hold for brain data. Furthermore, the assumptions underlying core SSL techniques may be mismatched with neural data properties. Techniques like Masked Autoencoders or BERT-style masking rely heavily on contextual redundancy and predictable structures (like grammar in language) to accurately reconstruct masked portions. While temporal dependencies exist in brain signals, they are often dynamic, non-stationary, and lack the rigid rules of syntax, potentially limiting the effectiveness of purely reconstruction-based objectives in capturing deep neurophysiological meaning. Simply masking random time points or channels might not encourage the model to learn the most relevant functional patterns. For contrastive learning, methods like SimCLR or MoCo typically assume that different augmented views of the same instance share core semantic identity (positive pairs) and that instances from different source samples are semantically distinct (negative pairs). This premise is challenged by brain data. High inter-subject variability means that neural responses from two different individuals performing the same task might be less similar than responses from the same individual across different tasks. Moreover, defining augmentations that preserve essential neural information while creating informative variations, without introducing artifacts, remains an open research question.

Consequently, while SSL frameworks offer powerful tools, their direct application may struggle to extract robust, transferable features from the inherently noisy and variable nature of brain signals without significant adaptation.

2. The gap between pre-training and true generalization:

Another critical observation is the current reliance of most foundation models on extensive, task-specific fine-tuning, which seems somewhat contrary to the ultimate goal of SSL—learning universal representations that generalize effectively across diverse downstream tasks with minimal labeled data. While SSL provides a powerful pre-training framework, the near-universal need for considerable fine-tuning suggests that the pre-trained representations, while beneficial, may not yet fully capture the kind of versatile, zero-shot or few-shot capabilities demonstrated by large language models like GPT in the NLP domain. Even NeuroLM [44], which represents a pioneering effort to mitigate fine-tuning through multi-task instruction tuning, shows improved adaptability but reportedly still lags behind task-specific fine-tuned models in performance benchmarks. This highlights that current SSL strategies for foundation models have yet to bridge the gap toward truly task-agnostic, general-purpose neural representations with zero-shot capabilities.

3. The need for domain-specific adaptation:

Finally, a significant limitation arises from the current tendency to adopt brain foundation models training strategies largely wholesale from established NLP and CV practices, often without deeply integrating domain-specific neuroscientific knowledge. Standard implementations of masked prediction and contrastive learning frequently overlook fundamental properties of brain function. First, functional connectivity, the brain operates as a network, yet masking or contrastive strategies rarely explicitly model or leverage the relationships between different brain regions or channels. Second, hierarchical processing, sensory and cognitive information is processed hierarchically, but standard SSL may not enforce or discover such hierarchies. Third, temporal dynamics & oscillations, brain activity unfolds dynamically over multiple timescales, involving critical oscillatory patterns that naive masking or instance discrimination might disrupt or fail to capture effectively.

Treating brain signals merely as generic time series or spatiotemporal data, without embedding constraints or objectives derived from neurobiology, limits the potential of these models. For instance, masking strategies could be made more neuro-plausible by considering known functional networks or temporal event structures, rather than random patches. Contrastive learning could be improved by defining positive pairs based on known neurophysiological states or by developing augmentation techniques grounded in understanding signal properties.

# Conclusion

While the adoption of SSL represents a significant advancement for analyzing large-scale neural datasets via brain foundation models, its current application faces substantial hurdles related to data characteristics, generalization limitations, and a lack of sufficient domain adaptation. The direct transfer of techniques successful in NLP and CV has yielded promising initial results but appears insufficient to fully unlock the potential of foundation models for neuroscience. Future progress likely hinges on developing novel SSL approaches specifically tailored to the unique properties and complexities of brain signals, integrating neurobiological insights more deeply into the learning objectives and architectures. Addressing these challenges is crucial for moving beyond direct adaptations towards large brain models that offer not only improved performance but also greater biological interpretability and genuine utility for advancing our understanding of the brain.

# Acknowledgment

# Declarations

Conflict of interest: The authors declare that they do not have any type of conflict of interest.

# References

[1] R. Balestriero *et al.*, "A cookbook of self-supervised learning," *arXiv Prepr. arXiv2304.12210*, 2023.

[2] P. Vincent, H. Larochelle, Y. Bengio, and P.-A. Manzagol, "Extracting and composing robust features with denoising autoencoders," in *Proceedings of the 25th international conference on Machine learning*, 2008, pp. 1096–1103.

[3] P. Vincent, H. Larochelle, I. Lajoie, Y. Bengio, P.-A. Manzagol, and L. Bottou, "Stacked denoising autoencoders: Learning useful representations in a deep network with a local denoising criterion.," *J. Mach. Learn. Res.*, vol. 11, no. 12, 2010.

[4] T. Mikolov, K. Chen, G. Corrado, and J. Dean, "Efficient estimation of word representations in vector space," *arXiv Prepr. arXiv1301.3781*, 2013.

[5] A. Vaswani *et al.*, "Attention is all you need," in *Advances in neural information processing systems*, 2017, pp. 5998–6008.

[6] H. Altaheri, G. Muhammad, and M. Alsulaiman, "Dynamic convolution with multilevel attention for EEG-based motor imagery decoding," *IEEE Internet Things J.*, p. 1, 2023.

[7] S. U. Amin, H. Altaheri, G. Muhammad, M. Alsulaiman, and A. Wadood, "Attention-Inception and Long Short-Term Memory-based Electroencephalography Classification for Motor Imagery Tasks in Rehabilitation," *IEEE Trans. Ind. Informatics*, vol. 18, no. 8, pp. 5412–5421, 2022.

[8] G. A. Altuwaijri, G. Muhammad, H. Altaheri, and M. Alsulaiman, "A Multi-Branch Convolutional Neural Network with Squeeze-and-Excitation Attention Blocks for EEG-Based Motor Imagery Signals Classification," *Diagnostics*, vol. 12, no. 4, pp. 1–16, 2022.

[9] H. Altaheri *et al.*, "Deep learning techniques for classification of electroencephalogram (EEG) motor imagery (MI) signals: a review," *Neural Comput. Appl.*, vol. 35, no. 20, pp. 14681–14722, 2023.

[10] W. B. Jiang, L. M. Zhao, and B. L. Lu, "Large Brain Model for Learning Generic Representations with Tremendous EEG Data in BCI," in *12th International Conference on Learning Representations, ICLR 2024*, 2024.

[11] J. del R. Millán *et al.*, "Combining brain–computer interfaces and assistive technologies: state-of-the-art and challenges," *Front. Neurosci.*, vol. 4, p. 161, 2010.

[12] L. J. Greenfield, J. D. Geyer, and P. R. Carney, *Reading EEGs: A practical approach*. Lippincott Williams & Wilkins, 2012.

[13] T. Ball, M. Kern, I. Mutschler, A. Aertsen, and A. Schulze-Bonhage, "Signal quality of simultaneously recorded invasive and non-invasive EEG," *Neuroimage*, vol. 46, no. 3, pp. 708–716, 2009.

[14] E. R. Kandel, J. H. Schwartz, T. M. Jessell, S. Siegelbaum, A. J. Hudspeth, and S. Mack, *Principles of neural science*, vol. 4.

McGraw-hill New York, 2000.
[15] "CHB-MIT Scalp EEG Database." [Online]. Available: https://archive.physionet.org/physiobank/charts/chbmit.png. [Accessed: 12-Apr-2020].
[16] K. He, X. Chen, S. Xie, Y. Li, P. Dollár, and R. Girshick, "Masked autoencoders are scalable vision learners," in *Proceedings of the IEEE/CVF conference on computer vision and pattern recognition*, 2022, pp. 16000–16009.
[17] R. Bommasani *et al.*, "On the opportunities and risks of foundation models," *arXiv Prepr. arXiv2108.07258*, 2021.
[18] W. Mai, J. Zhang, P. Fang, and Z. Zhang, "Brain-conditional multimodal synthesis: A survey and taxonomy," *IEEE Trans. Artif. Intell.*, 2024.
[19] D. Moher, A. Liberati, J. Tetzlaff, D. G. Altman, and P. Group, "Preferred reporting items for systematic reviews and meta-analyses: the PRISMA statement," *PLoS med*, vol. 6, no. 7, p. e1000097, 2009.
[20] R. Raina, A. Battle, H. Lee, B. Packer, and A. Y. Ng, "Self-taught learning: transfer learning from unlabeled data," in *Proceedings of the 24th international conference on Machine learning*, 2007, pp. 759–766.
[21] D. P. Kingma and M. Welling, "Auto-encoding variational bayes," *arXiv Prepr. arXiv1312.6114*, 2013.
[22] J. Pennington, R. Socher, and C. D. Manning, "Glove: Global vectors for word representation," in *Proceedings of the 2014 conference on empirical methods in natural language processing (EMNLP)*, 2014, pp. 1532–1543.
[23] C. Doersch, A. Gupta, and A. A. Efros, "Unsupervised visual representation learning by context prediction," in *Proceedings of the IEEE international conference on computer vision*, 2015, pp. 1422–1430.
[24] R. Zhang, P. Isola, and A. A. Efros, "Colorful image colorization," in *Computer Vision–ECCV 2016: 14th European Conference, Amsterdam, The Netherlands, October 11-14, 2016, Proceedings, Part III 14*, 2016, pp. 649–666.
[25] M. Noroozi and P. Favaro, "Unsupervised learning of visual representations by solving jigsaw puzzles," in *European conference on computer vision*, 2016, pp. 69–84.
[26] D. Pathak, P. Krahenbuhl, J. Donahue, T. Darrell, and A. A. Efros, "Context encoders: Feature learning by inpainting," in *Proceedings of the IEEE conference on computer vision and pattern recognition*, 2016, pp. 2536–2544.
[27] S. Gidaris, P. Singh, and N. Komodakis, "Unsupervised representation learning by predicting image rotations," *arXiv Prepr. arXiv1803.07728*, 2018.
[28] K. He, H. Fan, Y. Wu, S. Xie, and R. Girshick, "Momentum contrast for unsupervised visual representation learning," in *Proceedings of the IEEE/CVF conference on computer vision and pattern recognition*, 2020, pp. 9729–9738.
[29] T. Chen, S. Kornblith, M. Norouzi, and G. Hinton, "A simple framework for contrastive learning of visual representations," in *International conference on machine learning*, 2020, pp. 1597–1607.
[30] J.-B. Grill *et al.*, "Bootstrap your own latent-a new approach to self-supervised learning," *Adv. Neural Inf. Process. Syst.*, vol. 33, pp. 21271–21284, 2020.
[31] J. Devlin, M.-W. Chang, K. Lee, and K. Toutanova, "Bert: Pre-training of deep bidirectional transformers for language understanding," in *Proceedings of the 2019 conference of the North American chapter of the association for computational linguistics: human language technologies, volume 1 (long and short papers)*, 2019, pp. 4171–4186.
[32] A. Radford, K. Narasimhan, T. Salimans, and I. Sutskever, "Improving language understanding by generative pre-training," 2018.
[33] A. Radford, J. Wu, R. Child, D. Luan, D. Amodei, and I. Sutskever, "Language models are unsupervised multitask learners," *OpenAI blog*, vol. 1, no. 8, p. 9, 2019.
[34] W.-L. Zheng and B.-L. Lu, "Investigating critical frequency bands and channels for EEG-based emotion recognition with deep neural networks," *IEEE Trans. Auton. Ment. Dev.*, vol. 7, no. 3, pp. 162–175, 2015.
[35] S. Koelstra *et al.*, "Deap: A database for emotion analysis; using physiological signals," *IEEE Trans. Affect. Comput.*, vol. 3, no. 1, pp. 18–31, 2011.
[36] F. Setti *et al.*, "A modality-independent proto-organization of human multisensory areas," *Nat. Hum. Behav.*, vol. 7, no. 3, pp. 397–410, 2023.
[37] Z. Dong *et al.*, "Brain-JEPA: Brain dynamics foundation model with gradient positioning and spatiotemporal masking," *Adv. Neural Inf. Process. Syst.*, vol. 37, pp. 86048–86073, 2024.
[38] D. Z. Zhang, Z. Z. Yuan, Y. Yang, J. R. Chen, J. J. Wang, and Y. F. Li, "Brant: Foundation Model for Intracranial Neural Signal," *ADVANCES IN NEURAL INFORMATION PROCESSING SYSTEMS 36 (NEURIPS 2023)*, no. 37th Conference on Neural Information Processing Systems (NeurIPS). 2023.
[39] J. Wang *et al.*, "CBraMod: A Criss-Cross Brain Foundation Model for EEG Decoding," in *The Thirteenth International Conference on Learning Representations*, 2025.
[40] G. Wang, W. Liu, Y. He, C. Xu, L. Ma, and H. Li, "Eegpt: Pretrained transformer for universal and reliable representation of eeg signals," *Adv. Neural Inf. Process. Syst.*, vol. 37, pp. 39249–39280, 2024.
[41] Z. Xie *et al.*, "Simmim: A simple framework for masked image modeling," in *Proceedings of the IEEE/CVF conference on computer vision and pattern recognition*, 2022, pp. 9653–9663.
[42] Y. Chen *et al.*, "EEGFormer: Towards Transferable and Interpretable Large-Scale EEG Foundation Model," in *AAAI 2024 Spring Symposium on Clinical Foundation Models*, 2024.
[43] L. Wang, T. Suzumura, and H. Kanezashi, "GEFM: Graph-Enhanced EEG Foundation Model," *arXiv.org*, 2024.
[44] W. Jiang, Y. Wang, B. Lu, and D. Li, "NeuroLM: A Universal Multi-task Foundation Model for Bridging the Gap between Language and EEG Signals," in *The Thirteenth International Conference on Learning Representations*, 2025.
[45] C. Fang *et al.*, "Promoting Cross-Modal Representations to Improve Multimodal Foundation Models for Physiological Signals," in *NeurIPS*, 2024.
[46] A. Kommineni, K. Avramidis, R. Leahy, and S. Narayanan, "Knowledge-guided EEG Representation Learning," in *46th Annual International Conference of the IEEE Engineering in Medicine and Biology Society (EMBC)*, 2024, pp. 1–6.
[47] W. Cui *et al.*, "Neuro-GPT: Towards A Foundation Model For EEG," *Proceedings - International Symposium on Biomedical Imaging*, no. 21st IEEE International Symposium on Biomedical Imaging (ISBI). 2024.
[48] T. Yue *et al.*, "EEGPT: Unleashing the Potential of EEG Generalist Foundation Model by Autoregressive Pre-training," *arXiv Prepr. arXiv2410.19779*, 2024.
[49] J. Wang, Z. Song, Z. Ma, X. Qiu, M. Zhang, and Z. Zhang, "Enhancing EEG-to-Text Decoding through Transferable Representations from Pre-trained Contrastive EEG-Text Masked Autoencoder," *PROCEEDINGS OF THE 62ND ANNUAL MEETING OF THE ASSOCIATION FOR COMPUTATIONAL LINGUISTICS, VOL 1: LONG PAPERS*, no. 62nd Annual Meeting of the Association-for-Computational-Linguistics (ACL) / Student Research Workshop (SRW). pp. 7278–7292, 2024.
[50] D. Wu, S. Li, J. Yang, and M. Sawan, "Neuro-BERT: Rethinking Masked Autoencoding for Self-Supervised Neurological Pretraining," *IEEE J. Biomed. Heal. Informatics*, 2024.
[51] C. Wang *et al.*, "Brainbert: Self-Supervised Representation Learning for Intracranial Recordings," *11th Int. Conf. Learn. Represent. ICLR 2023*, 2023.
[52] A. van den Oord, Y. Li, and O. Vinyals, "Representation learning with contrastive predictive coding," *arXiv Prepr. arXiv1807.03748*, 2018.
[53] X. Chen and K. He, "Exploring simple siamese representation learning," in *Proceedings of the IEEE/CVF conference on computer vision and pattern recognition*, 2021, pp. 15750–15758.
[54] M. Caron *et al.*, "Emerging properties in self-supervised vision transformers," in *Proceedings of the IEEE/CVF international conference on computer vision*, 2021, pp. 9650–9660.
[55] J. Zbontar, L. Jing, I. Misra, Y. LeCun, and S. Deny, "Barlow twins: Self-supervised learning via redundancy reduction," in *International conference on machine learning*, 2021, pp. 12310–12320.
[56] A. Bardes, J. Ponce, and Y. Lecun, "VICReg: Variance-Invariance-Covariance Regularization For Self-Supervised Learning," in *ICLR 2022-International Conference on Learning Representations*, 2022.
[57] D. H. Cai, J. R. Chen, Y. Yang, T. Liu, Y. F. Li, and ACM, "MBrain: A Multi-channel Self-Supervised Learning Framework for Brain Signals," *PROCEEDINGS OF THE 29TH ACM SIGKDD CONFERENCE ON KNOWLEDGE DISCOVERY AND DATA MINING, KDD 2023*, no. 29th ACM SIGKDD Conference on Knowledge Discovery and Data Mining (KDD). pp. 130–141,

2023.

[58] D. Kostas, S. Aroca-Ouellette, and F. Rudzicz, "BENDR: Using Transformers and a Contrastive Self-Supervised Learning Task to Learn From Massive Amounts of EEG Data," *Front. Hum. Neurosci.*, vol. 15, 2021.

[59] A. Baevski, Y. Zhou, A. Mohamed, and M. Auli, "wav2vec 2.0: A framework for self-supervised learning of speech representations," *Adv. Neural Inf. Process. Syst.*, vol. 33, pp. 12449–12460, 2020.

[60] W. Li *et al.*, "Self-supervised contrastive learning for EEG-based cross-subject motor imagery recognition," *J. Neural Eng.*, vol. 21, no. 2, 2024.

[61] C. Yang, M. Westover, and J. Sun, "Biot: Biosignal transformer for cross-data learning in the wild," *Adv. Neural Inf. Process. Syst.*, vol. 36, pp. 78240–78260, 2023.

[62] S. Gijsen and K. Ritter, "EEG-Language Modeling for Pathology Detection," *arXiv Prepr. arXiv2409.07480*, 2024.

[63] A. Radford *et al.*, "Learning transferable visual models from natural language supervision," in *International conference on machine learning*, 2021, pp. 8748–8763.

[64] H. Xu *et al.*, "Videoclip: Contrastive pre-training for zero-shot video-text understanding," *arXiv Prepr. arXiv2109.14084*, 2021.

[65] S. Ma, Z. Zeng, D. McDuff, and Y. Song, "Active contrastive learning of audio-visual video representations," *arXiv Prepr. arXiv2009.09805*, 2020.

[66] R. Ye, M. Wang, and L. Li, "Cross-modal contrastive learning for speech translation," *arXiv Prepr. arXiv2205.02444*, 2022.

[67] X. Wei *et al.*, "Multi-modal cross-domain self-supervised pre-training for fMRI and EEG fusion," *Neural Networks*, vol. 184, 2025.

[68] D. Z. Zhang, Z. Z. Yuan, J. R. Chen, K. R. Chen, Y. Yang, and A. C. MACHINERY, "Brant-X: A Unified Physiological Signal Alignment Framework," *PROCEEDINGS OF THE 30TH ACM SIGKDD CONFERENCE ON KNOWLEDGE DISCOVERY AND DATA MINING, KDD 2024*, no. 30th ACM SIGKDD Conference on Knowledge Discovery and Data Mining. pp. 4155–4166, 2024.

[69] Z. S. Ye, L. A. Yao, Y. Zhang, and S. Gustin, "Self-supervised cross-modal visual retrieval from brain activities," *PATTERN Recognit.*, vol. 145, 2024.

[70] J. Zhou, Y. Duan, F. Chang, T. Do, Y.-K. Wang, and C.-T. Lin, "BELT-2: Bootstrapping EEG-to-Language representation alignment for multi-task brain decoding," *arXiv Prepr. arXiv2409.00121*, 2024.

[71] A. Akbarinia, "Optimising EEG decoding with refined sampling and multimodal feature integration," *arXiv Prepr. arXiv2409.20086*, 2024.

[72] Y. Song, B. Liu, X. Li, N. Shi, Y. Wang, and X. Gao, "Decoding Natural Images from EEG for Object Recognition," in *The Twelfth International Conference on Learning Representations*, 2024.

[73] F. Yang *et al.*, "NeuroBind: Towards Unified Multimodal Representations for Neural Signals," *arXiv.org*, 2024.

[74] W. Weng, Y. Gu, Q. Zhang, Y. Huang, C. Miao, and Y. Chen, "A Knowledge-Driven Cross-view Contrastive Learning for EEG Representation," *arXiv.org*, 2023.

[75] H. Banville, O. Chehab, A. Hyvärinen, D.-A. Engemann, and A. Gramfort, "Uncovering the structure of clinical EEG signals with self-supervised learning," *J. Neural Eng.*, vol. 18, no. 4, p. 46020, 2021.

[76] X. Kong, K. Huang, P. Li, and L. Zhang, "Toward Generalizing Visual Brain Decoding to Unseen Subjects," in *The Thirteenth International Conference on Learning Representations*, 2025.

[77] T. Brown *et al.*, "Language models are few-shot learners," *Adv. Neural Inf. Process. Syst.*, vol. 33, pp. 1877–1901, 2020.

[78] S. Bubeck *et al.*, "Sparks of Artificial General Intelligence: Early experiments with GPT-4." ArXiv, 2023.

[79] Z. Yuan, D. Zhang, J. Chen, G. Gu, and Y. Yang, "Brant-2: Foundation model for brain signals," *arXiv e-prints*, p. arXiv-2402, 2024.

[80] Z. Yuan, F. Shen, M. Li, Y. Yu, C. Tan, and Y. Yang, "BrainWave: A Brain Signal Foundation Model for Clinical Applications," *arXiv Prepr. arXiv2402.10251*, 2024.

[81] H.-Y. S. Chien, H. Goh, C. M. Sandino, and J. Y. Cheng, "{MAEEG}: Masked Auto-encoder for {EEG} Representation Learning," in *NeurIPS 2022 Workshop on Learning from Time Series for Health*, 2022.

[82] C. A. Barbano, M. Brunello, B. Dufumier, and M. Grangetto, "Anatomical Foundation Models for Brain MRIs," *arXiv Prepr. arXiv2408.07079*, 2024.

[83] J. O. Caro *et al.*, "BrainLM: A foundation model for brain activity recordings," in *The Twelfth International Conference on Learning Representations*, 2024.

[84] Y. Yang, Y. Mao, X. Liu, and X. Liu, "BrainMAE: A Region-aware Self-supervised Learning Framework for Brain Signals," *arXiv Prepr. arXiv2406.17086*, 2024.

[85] J. Cox *et al.*, "BrainSegFounder: towards 3D foundation models for neuroimage segmentation," *Med. Image Anal.*, vol. 97, p. 103301, 2024.

[86] A. Dimofte *et al.*, "CEReBrO: Compact Encoder for Representations of Brain Oscillations Using Efficient Alternating Attention," *arXiv Prepr. arXiv2501.10885*, 2025.

[87] Z. He *et al.*, "FM-APP: Foundation Model for Any Phenotype Prediction via fMRI to sMRI Knowledge Transfer," *IEEE Trans. Med. Imaging*, 2024.

[88] D. Karimi, "An Approach to Building Foundation Models for Brain Image Analysis," in *International Conference on Medical Image Computing and Computer-Assisted Intervention*, 2024, pp. 421–431.

[89] E. Shi *et al.*, "FoME: A Foundation Model for EEG using Adaptive Temporal-Lateral Attention Scaling," *arXiv Prepr. arXiv2409.12454*, 2024.

[90] C.-S. Chen, Y.-J. Chen, and A. H.-W. Tsai, "Large Cognition Model: Towards Pretrained EEG Foundation Model," *arXiv Prepr. arXiv2502.17464*, 2025.

[91] E. Shi *et al.*, "MEET: A Multi-Band EEG Transformer for Brain States Decoding," *IEEE Trans. Biomed. Eng.*, vol. 71, no. 5, pp. 1442–1453, 2024.

[92] W. Cui, H. Akrami, G. Zhao, A. A. Joshi, and R. M. Leahy, "Meta Transfer of Self-Supervised Knowledge: Foundation Model in Action for Post-Traumatic Epilepsy Prediction," *ArXiv*, p. arXiv-2312, 2023.

[93] S. Sihag, G. Mateos, and A. Ribeiro, "Towards a foundation model for brain age prediction using covariance neural networks," *arXiv Prepr. arXiv2402.07684*, 2024.

[94] X. Zhang, Z. Zhao, T. Tsiligkaridis, and M. Zitnik, "Self-supervised contrastive pre-training for time series via time-frequency consistency," *Adv. Neural Inf. Process. Syst.*, vol. 35, pp. 3988–4003, 2022.

[95] H. Altaheri, G. Muhammad, and M. Alsulaiman, "Physics-Informed Attention Temporal Convolutional Network for EEG-Based Motor Imagery Classification," *IEEE Trans. Ind. Informatics*, vol. 19, no. 2, pp. 2249–2258, 2023.

[96] S. U. Amin, H. Altaheri, G. Muhammad, M. Alsulaiman, and W. Abdul, "Attention based Inception model for robust EEG motor imagery classification," in *2021 IEEE International Instrumentation and Measurement Technology Conference (I2MTC)*, 2021, pp. 1–6.

[97] V. Jayaram and A. Barachant, "MOABB: trustworthy algorithm benchmarking for BCIs," *J. Neural Eng.*, vol. 15, no. 6, p. 66011, 2018.

[98] H. Altaheri, G. Muhammed, S. U. Amin, and M. Alsulaiman, "REH-MI: EEG Motor Imagery Dataset from the Same Limb for Rehabilitation Applications." IEEE Dataport, 2025.

[99] Y. Li *et al.*, "Neurobolt: Resting-state eeg-to-fmri synthesis with multi-dimensional feature mapping," *Adv. Neural Inf. Process. Syst.*, vol. 37, pp. 23378–23405, 2024.

[100] W. Yao, Z. Lyu, M. Mahmud, N. Zhong, B. Lei, and S. Wang, "CATD: Unified Representation Learning for EEG-To-fMRI Cross-Modal Generation," *IEEE Trans. Med. Imaging*, 2025.

[101] M. Ferrante, T. Boccato, and N. Toschi, "Towards neural foundation models for vision: Aligning eeg, meg and fmri representations to perform decoding, encoding and modality conversion," in *ICLR 2024 Workshop on Representational Alignment*, 2024.

[102] H. Amrani, D. Micucci, and P. Napoletano, "Deep Representation Learning for Open Vocabulary Electroencephalograph y-to-Text Decoding," *IEEE J. Biomed. Heal. Informatics*, pp. 1–12, 2024.

[103] L. Yifei, Y. Hengwei, and L. Shuhang, "LLMs Help Alleviate the Cross-Subject Variability in Brain Signal and Language Alignment," 2025.

[104] S. Palazzo, C. Spampinato, I. Kavasidis, D. Giordano, J. Schmidt, and M. Shah, "Decoding Brain Representations by Multimodal Learning of Neural Activity and Visual Features," *IEEE Trans. Pattern Anal. Mach. Intell.*, vol. 43, no. 11, pp. 3833–3849, 2021.

[105] C. Du, K. Fu, J. Li, and H. He, "Decoding visual neural representations by multimodal learning of brain-visual-linguistic features," *IEEE Trans. Pattern Anal. Mach. Intell.*, vol. 45, no. 9, pp. 10760–10777, 2023.

[106] Y. Bai, X. Wang, Y.-P. Cao, Y. Ge, C. Yuan, and Y. Shan, "DreamDiffusion: High-Quality EEG-to-Image Generation with Temporal Masked Signal Modeling and CLIP Alignment," in *European Conference on Computer Vision*, 2024, pp. 472–488.

[107] Y.-T. Lan *et al.*, "Seeing through the brain: image reconstruction of visual perception from human brain signals," *arXiv Prepr. arXiv2308.02510*, 2023.

[108] J. Li, Z. Song, J. Wang, M. Zhang, and Z. Zhang, "BrainECHO: Semantic Brain Signal Decoding through Vector-Quantized Spectrogram Reconstruction for Whisper-Enhanced Text Generation," *arXiv Prepr. arXiv2410.14971*, 2024.

[109] W. X. Lei *et al.*, "VIT-LENS: Towards Omni-modal Representations," *2024 IEEE/CVF CONFERENCE ON COMPUTER VISION AND PATTERN RECOGNITION (CVPR)*, no. IEEE/CVF Conference on Computer Vision and Pattern Recognition (CVPR). pp. 26637–26647, 2024.

[110] C. Spampinato, S. Palazzo, I. Kavasidis, D. Giordano, N. Souly, and M. Shah, "Deep learning human mind for automated visual classification," in *Proceedings of the IEEE conference on computer vision and pattern recognition*, 2017, pp. 6809–6817.

[111] Y. Guo, T. Liu, X. Zhang, A. Wang, and W. Wang, "End-to-end translation of human neural activity to speech with a dual–dual generative adversarial network," *Knowledge-based Syst.*, vol. 277, p. 110837, 2023.

[112] I. Daly *et al.*, "Neural and physiological data from participants listening to affective music," *Sci. data*, vol. 7, no. 1, p. 177, 2020.

[113] H. Chen, O. Gaoxiang, and X. Li, "Extracting Temporal-Spectral-Spatial Representation of EEG Using Self-Supervised Learning for the Identification of Children with ASD," in *2023 IEEE 13th International Conference on CYBER Technology in Automation, Control, and Intelligent Systems (CYBER)*, 2023, pp. 1263–1266.

[114] L. Bollens, B. Accou, M. Gillis, W. Verheijen, and T. Francart, "SparrKULee: A speech-evoked auditory response repository of the KU Leuven, containing EEG of 85 participants," 2023.

[115] R. Peng *et al.*, "TIE-EEGNet: Temporal information enhanced EEGNet for seizure subtype classification," *IEEE Trans. Neural Syst. Rehabil. Eng.*, vol. 30, pp. 2567–2576, 2022.

[116] A. T. Gifford, K. Dwivedi, G. Roig, and R. M. Cichy, "A large and rich EEG dataset for modeling human visual object recognition," *Neuroimage*, vol. 264, p. 119754, 2022.

[117] A. Babayan *et al.*, "A mind-brain-body dataset of MRI, EEG, cognition, emotion, and peripheral physiology in young and old adults," *Sci. data*, vol. 6, no. 1, pp. 1–21, 2019.

[118] M.-H. Lee *et al.*, "EEG dataset and OpenBMI toolbox for three BCI paradigms: an investigation into BCI illiteracy," *Gigascience*, vol. 8, no. 5, p. giz002, 2019.

[119] T. Song, W. Zheng, C. Lu, Y. Zong, X. Zhang, and Z. Cui, "MPED: A multi-modal physiological emotion database for discrete emotion recognition," *IEEE Access*, vol. 7, pp. 12177–12191, 2019.

[120] M. M. Ghassemi *et al.*, "You snooze, you win: the physionet/computing in cardiology challenge 2018," in *2018 Computing in Cardiology Conference (CinC)*, 2018, vol. 45, pp. 1–4.

[121] W.-L. Zheng, W. Liu, Y. Lu, B.-L. Lu, and A. Cichocki, "Emotionmeter: A multimodal framework for recognizing human emotions," *IEEE Trans. Cybern.*, vol. 49, no. 3, pp. 1110–1122, 2018.

[122] S. López, I. Obeid, and J. Picone, "Automated interpretation of abnormal adult electroencephalograms," *MS Thesis, Temple Univ.*, 2017.

[123] S. Katsigiannis and N. Ramzan, "DREAMER: A database for emotion recognition through EEG and ECG signals from wireless low-cost off-the-shelf devices," *IEEE J. Biomed. Heal. informatics*, vol. 22, no. 1, pp. 98–107, 2017.

[124] R.-N. Duan, J.-Y. Zhu, and B.-L. Lu, "Differential entropy feature for EEG-based emotion classification," in *2013 6th international IEEE/EMBS conference on neural engineering (NER)*, 2013, pp. 81–84.

[125] A. Temko, A. Sarkar, and G. Lightbody, "Detection of seizures in intracranial EEG: UPenn and Mayo Clinic's seizure detection challenge," in *2015 37th Annual International Conference of the IEEE Engineering in Medicine and Biology Society (EMBC)*, 2015, pp. 6582–6585.

[126] C. O'reilly, N. Gosselin, J. Carrier, and T. Nielsen, "Montreal Archive of Sleep Studies: an open-access resource for instrument benchmarking and exploratory research," *J. Sleep Res.*, vol. 23, no. 6, pp. 628–635, 2014.

[127] M. Tangermann *et al.*, "Review of the BCI competition IV," *Front. Neurosci.*, vol. 6, p. 55, 2012.

[128] M. Soleymani, J. Lichtenauer, T. Pun, and M. Pantic, "A multimodal database for affect recognition and implicit tagging," *IEEE Trans. Affect. Comput.*, vol. 3, no. 1, pp. 42–55, 2011.

[129] A. H. Shoeb, "Application of machine learning to epileptic seizure onset detection and treatment." Massachusetts Institute of Technology, 2009.

[130] G. Schalk, D. J. McFarland, T. Hinterberger, N. Birbaumer, and J. R. Wolpaw, "BCI2000: a general-purpose brain-computer interface (BCI) system," *IEEE Trans. Biomed. Eng.*, vol. 51, no. 6, pp. 1034–1043, 2004.

[131] R. G. Andrzejak, K. Lehnertz, F. Mormann, C. Rieke, P. David, and C. E. Elger, "Indications of nonlinear deterministic and finite-dimensional structures in time series of brain electrical activity: Dependence on recording region and brain state," *Phys. Rev. E*, vol. 64, no. 6, p. 61907, 2001.

[132] M. Ogg and W. G. Coon, "Self-Supervised Transformer Model Training for a Sleep-EEG Foundation Model," *bioRxiv*, 2024.

[133] A. Défossez, C. Caucheteux, J. Rapin, O. Kabeli, and J.-R. King, "Decoding speech perception from non-invasive brain recordings," *Nat. Mach. Intell.*, vol. 5, no. 10, pp. 1097–1107, 2023.

[134] H. Zheng *et al.*, "Learning from models beyond fine-tuning," *Nat. Mach. Intell.*, vol. 7, no. 1, pp. 6–17, 2025.

[135] J. Lévy *et al.*, "Brain-to-Text Decoding: A Non-invasive Approach via Typing," *Meta-Research*, 2025.

[136] M. Brickwedde *et al.*, "Applications of OPM-MEG for translational neuroscience: a perspective," *Transl. Psychiatry*, vol. 14, no. 1, p. 341, 2024.

[137] D. Li, C. Wei, S. Li, J. Zou, H. Qin, and Q. Liu, "Visual decoding and reconstruction via eeg embeddings with guided diffusion," *arXiv Prepr. arXiv2403.07721*, 2024.

[138] H. Jo, Y. Yang, J. Han, Y. Duan, H. Xiong, and W. H. Lee, "Are eeg-to-text models working?," *arXiv Prepr. arXiv2405.06459*, 2024.

[139] Z. Wang and H. Ji, "Open vocabulary electroencephalography-to-text decoding and zero-shot sentiment classification," in *Proceedings of the AAAI Conference on Artificial Intelligence*, 2022, vol. 36, no. 5, pp. 5350–5358.

[140] J. Z. Zhou, Y. Q. Duan, Y. C. Chang, Y. K. Wang, and C. T. Lin, "BELT: Bootstrapped EEG-to-Language Training by Natural Language Supervision," *IEEE Trans. NEURAL Syst. Rehabil. Eng.*, vol. 32, pp. 3278–3288, 2024.

[141] X. C. Feng, X. C. Feng, B. Qin, and T. Liu, "Aligning Semantic in Brain and Language: A Curriculum Contrastive Method for Electroencephalography-to-Text Generation," *IEEE Trans. NEURAL Syst. Rehabil. Eng.*, vol. 31, pp. 3874–3883, 2023.

[142] N. Xi, S. Zhao, H. Wang, C. Liu, B. Qin, and T. Liu, "UniCoRN: Unified cognitive signal reconstruction bridging cognitive signals and human language," in *Proceedings of the 61st Annual Meeting of the Association for Computational Linguistics (Volume 1: Long Papers)*, 2023, pp. 13277–13291.

[143] X. Feng, X. Feng, and B. Qin, "Semantic-aware contrastive learning for electroencephalography-to-text generation with curriculum learning," *arXiv Prepr. arXiv2301.09237*, 2023.

[144] Y. Q. Duan, J. Z. Zhou, Z. Wang, Y. K. Wang, and C. T. Lin, "DeWave: Discrete EEG Waves Encoding for Brain Dynamics to Text Translation," *ADVANCES IN NEURAL INFORMATION PROCESSING SYSTEMS 36 (NEURIPS 2023)*, no. 37th Conference on Neural Information Processing Systems (NeurIPS). 2023.